\documentclass[preprint,11pt,times,authoryear,nopreprintline]{elsarticle}

\usepackage[
    a4paper,
    left=2.2cm,
    right=2.2cm,
    top=2.5cm,
    bottom=2.5cm
]{geometry}
\usepackage{longtable}
\usepackage{booktabs}
\usepackage{tabularx}
\usepackage{array}
\usepackage[T1]{fontenc}
\usepackage[utf8]{inputenc}
\usepackage{amsmath,amssymb}
\usepackage{booktabs}
\usepackage{multirow}
\usepackage{array}
\usepackage{tabularx}
\usepackage{adjustbox}
\usepackage{graphicx}
\usepackage{microtype}
\usepackage{placeins}
\usepackage{float}
\usepackage{enumitem}
\usepackage{url}
\usepackage{caption}
\usepackage{afterpage}
\usepackage[
    colorlinks=true,
    citecolor=blue,
    linkcolor=blue,
    urlcolor=blue
]{hyperref}

\setlist[itemize]{
    leftmargin=1.6em,
    labelsep=0.5em,
    itemindent=0pt,
    topsep=3pt,
    itemsep=2pt,
    parsep=0pt
}
\setlist[enumerate]{leftmargin=*}

\journal{Transportation Research Part C: Emerging Technologies}
\biboptions{authoryear,round}

\begin{document}
\begin{frontmatter}

\title{Agentic discovery of interpretable car-following law 
from multi-source trajectory data with a knowledge-driven framework}

\author[aff1]{Jiamei Xu}
\author[aff1]{Jie Sun\corref{cor1}}
% \author[aff2]{Daiheng Ni}
\author[aff1]{Keshuang Tang}
\author[aff1]{Jian Sun}

\affiliation[aff1]{
    organization={College of Transportation Engineering \& Key Laboratory of Road and Traffic Engineering of the Ministry of Education, Tongji University},
    city={Shanghai},
    postcode={201804},
    country={China}
}

% \affiliation[aff2]{
%     organization={Department of Civil and Environmental Engineering, University of Massachusetts Amherst},
%     city={Amherst, MA},
%     postcode={01003},
%     country={USA}
% }

\cortext[cor1]{%
\begin{tabular}[t]{@{}l@{}}
Corresponding author.\\
E-mail addresses: jie\_sun@tongji.edu.cn (J. Sun).
\end{tabular}%
}

\begin{abstract}
Car-following (CF) modeling is a fundamental component of traffic flow theory and underpins the analysis of collective traffic dynamics, while its modeling becomes more challenging with the increasing presence of automated vehicles. Analytical CF models offer interpretable representations of longitudinal vehicle interactions, but their prescribed functional structures limit the generalizability in capturing diverse driving behavior. Discovering interpretable CF laws from trajectory data has been pursued with symbolic regression and, more recently, with LLM agents that automate equation search. Yet most existing methods evaluate candidate laws primarily by how accurately they reconstruct the trajectories used for calibration. Whether a discovered equation captures interaction patterns that generalize across heterogeneous driving environments therefore remains unclear. In this study, we present a knowledge-driven agentic framework that draws on classical CF formulations as domain references, combines agentic hypothesis generation with state-dependent error diagnosis, and evaluates candidate laws in terms of cross-source applicability and dynamical validity rather than fitting accuracy alone. Using trajectory data from NGSIM US-101, highD, and exiD, the framework identifies an explicit CF law, the Adaptive Equilibrium Response Model (AERM), which couples state-dependent relative-speed response with spacing-dependent equilibrium-speed restoration. AERM admits analytical equilibrium relationships in closed form, including a fundamental diagram whose critical speed, critical density and capacity are all determined by a single dimensionless group, together with explicit local and string stability conditions. Evaluated against seven classical and four agent-based CF models on three external datasets (Shanghai naturalistic driving data, NGSIM I-80, and Waymo dataset), AERM achieves the lowest calibration error under trajectory-specific calibration, reducing spacing RMSE by 9.11\%, 7.59\%, and 21.73\% relative to the best-performing baselines. These results demonstrate the fitting flexibility and analytical tractability of the discovered CF law, and identify the conditions under which microscopic parameter heterogeneity, rather than functional form, governs calibration accuracy.

\end{abstract}


\end{frontmatter}

% =========================================================
% 1. Introduction — revised for the 15 October 2026 AERM study
% =========================================================
\section{Introduction}
\label{sec:introduction}

Car-following (CF) models describe how a following vehicle adjusts its longitudinal motion in response to its leader and the prevailing traffic state. As a fundamental component of traffic flow theory, they determine how individual trajectories unfold and how disturbances form and propagate along a traffic stream \citep{Brackstone1999,Treiber2000,Jiang2001, Saifuzzaman2014}. Driving behavior varies substantially across drivers, road types, and traffic conditions, and the growing deployment of automated vehicles further amplifies such diversity, since their longitudinal control is implemented via engineered controllers whose response characteristics deviate systematically from human driving patterns. This inherent variability complicates the representation of longitudinal vehicle interactions using a single closed-form equation. A viable CF law should therefore capture nonlinear, state-contingent response patterns while retaining sufficient mathematical tractability to support rigorous analysis of equilibrium states, fundamental diagrams, and traffic flow stability.

Classical CF research has established interpretable mathematical models based on different behavioral assumptions. For instance, stimulus-response models describe acceleration through relative motion \citep{Gazis1961,Helly1959}; Gipps' model incorporates safety-oriented speed constraints \citep{Gipps1981}; the optimal velocity model (OVM) describes relaxation toward a spacing-dependent desired speed \citep{Bando1995}; Newell's model characterizes vehicle motion through spacing-dependent speed relationships and trajectory translation \citep{Newell2002}; the full velocity difference model (FVDM) introduces relative-speed feedback \citep{Jiang2001}; and the intelligent driver model (IDM) combines free-flow acceleration with desired-spacing interactions \citep{Treiber2000}. Their development generally follows a \emph{structure-first} paradigm, in which behavioral assumptions determine the mathematical form and empirical observations are subsequently used to calibrate model parameters \citep{Kesting2008}. Nevertheless, these formulations provide explicit connections between microscopic driving behavior and traffic flow theory, which embody accumulated knowledge of longitudinal driving behavior, including behavioral assumptions, governing relationships, and the physical interpretations of model parameters. Extensions with heterogeneous parameters or additional response terms increase flexibility, yet the represented interactions remain constrained by the prescribed structures \citep{Ossen2011,Saifuzzaman2014}. Consequently, although classical CF models provide valuable theoretical knowledge and interpretable behavioral descriptions, their predefined formulations may not fully capture the diverse interactions present in heterogeneous driving data. This motivates the discovery of analytical CF relationships directly from empirical trajectories, drawing on existing traffic flow theory while allowing the governing mathematical structures to emerge from observed driving behavior. 

High-resolution trajectory data and advances in machine learning have enabled data-driven approaches to CF modeling. Neural and nonparametric models can approximate complex driving responses, while their implicit representations require additional techniques for extracting explicit mechanisms and analyzing traffic flow dynamics \citep{Papathanasopoulou2015,Huang2018,Zhang2024NNStability}. Symbolic regression provides a complementary approach by discovering analytical expressions from observational data. Within traffic research, SciNet-CFM combines neural representation learning with symbolic regression to identify optimal traffic state relationships within analytical CF frameworks \citep{SciNetCFM2025}. SR-LLM integrates retrieval-augmented domain knowledge with reinforcement-learning-based symbolic regression to discover explicit acceleration equations from vehicle trajectories \citep{SRLLM2025}. Related approaches have employed deep symbolic regression to identify CF dynamics \citep{angah2024discovering}, while NOVA has investigated the recurrence and transferability of discovered mathematical structures across freeway observations \citep{NOVA2026}. These studies demonstrate the feasibility of discovering interpretable CF relationships from empirical trajectory data.

Recent advances in large language models (LLMs) and scientific agents have further accelerated autonomous equation discovery. LaSR exploits reusable concepts extracted from high-performing equations to guide evolutionary exploration \citep{LaSR2024}, while LLM-SR combines scientific priors with program-based equation generation and iterative optimization \citep{LLMSR2025}. SR-Scientist extends this process by enabling agents to analyze data, evaluate candidate equations, and refine hypotheses through tool-based interaction \citep{SRScientist2026}. KeplerAgent further incorporates intermediate physical reasoning to guide symbolic search \citep{KeplerAgent2026}. In transportation, TR-Agent demonstrates autonomous refinement of established traffic models through knowledge-guided hypothesis generation, implementation, and evaluation \citep{TRAgent2025}. Taken together, these advances have substantially improved the efficiency and autonomy of mathematical model discovery, creating new opportunities to investigate the behavioral and physical significance of the resulting equations.

Despite this progress, several challenges remain in discovering analytical CF laws from heterogeneous trajectory data. Although classical CF models provide valuable domain knowledge, it remains unclear how their governing formulations and physical interpretations can be harnessed to guide agentic hypothesis generation without unduly constraining the discovery of novel mathematical structures. First, evaluation is overwhelmingly dominated by goodness-of-fit metrics, as candidate equations are ranked by reconstruction error on calibration data, which measures the flexibility of a functional form rather than the validity of a behavioral law. Empirical CF data exhibit substantial heterogeneity in driving styles, traffic states, and observation conditions, potentially biasing equation discovery toward dataset-specific characteristics and limiting the applicability of the resulting structures to unseen driving environments. 
Additionally, physical plausibility is commonly enforced either by a small set of binary screens or by criteria that depend on a natural-language justification, and in the latter case the same agent that proposed a structure also supplies the text against which the structure is scored.
Identifying generic analytical CF laws for heterogeneous trajectory sources and systematically assessing their behavioral fidelity and traffic flow dynamics validity remain missing in the literature.
%These challenges motivate an diagnosis-guided discovery process that integrates state-dependent structural exploration with physics-aware evaluation, while emphasizing the cross-source applicability and dynamical validity of the resulting CF laws.

To address these limitations, we develop a knowledge-driven agentic framework for discovering interpretable CF laws. The framework provides the reasoning agent with the governing formulations and behavioral assumptions of classical CF models as domain priors. Guided by this prior knowledge together with diagnosed state-dependent reconstruction residuals, the agent constructs and iteratively revises structural hypotheses, while a complementary evolutionary symbolic search explores alternative mathematical formulations. Candidate equations are assessed through a physics-aware multi-objective criterion that jointly considers reconstruction accuracy, physical acceptability, and interpretability. Applied to NGSIM US-101, highD, and exiD datasets, the framework identifies an explicit CF law, termed the Adaptive Equilibrium Response Model (\textbf{AERM}), which combines state-dependent relative-speed response with nonlinear restoration toward an equilibrium speed. Its analytical structure enables the derivation of equilibrium relationships, a fundamental diagram in closed form, and explicit local and string stability conditions. We further evaluate AERM against eleven established CF models on three external datasets (Shanghai naturalistic driving data (NDS), NGSIM I-80, and Waymo datasets) under both dataset-level and trajectory-specific parameter calibration, and investigate its dynamical properties through equilibrium, stability, and platoon disturbance analyses.
The main contributions are summarized as follows:

\begin{itemize}
    \item \textbf{A knowledge-driven agentic framework for CF law discovery.} We develop an iterative loop in which an agent draws on classical CF formulations and their behavioral assumptions as domain references, combining this knowledge with persistent, state-dependent reconstruction errors to formulate testable structural hypotheses. %Evolutionary search explores the resulting candidates, and physics-aware multi-objective evaluation selects and refines them, with candidate laws verified across heterogeneous trajectory sources without prescribing the final equation in advance.
    \item \textbf{An interpretable empirical CF law discovered.} We identify AERM, an explicit CF law with state-dependent relative-speed sensitivity and nonlinear equilibrium-speed restoration, whose analytical structure allows the derivation of steady-state relationships, the associated fundamental diagram, and traffic flow stability. 
    \item \textbf{Comprehensive behavioral and dynamical evaluation.} We assess the discovered CF law against seven classical CF models and four agent-based CF models using real-world trajectory calibration under both common- and trajectory-specific parameter calibration, complemented by equilibrium analysis, stability analysis, and simulations of its dynamical properties.
\end{itemize}

% =========================================================
% 2. Related work
% =========================================================

\section{Related Work}
\label{sec:related_work}

\subsection{Analytical CF Models}

Classical analytical CF models describe longitudinal vehicle interactions through explicit mathematical formulations grounded in behavioral assumptions and traffic flow theory. Representative models include the GHR stimulus-response family \citep{Gazis1961}, the safety-based Gipps model \citep{Gipps1981}, Newell's trajectory-translation model \citep{Newell2002}, the Optimal Velocity Model (OVM) \citep{Bando1995}, the Full Velocity Difference Model (FVDM) \citep{Jiang2001}, and the Intelligent Driver Model (IDM) \citep{Treiber2000}. These formulations characterize different aspects of longitudinal driving behavior, including responses to relative motion, desired-speed adaptation, and spacing regulation, while providing a mathematical basis for equilibrium and stability analysis.

Subsequent studies have sought to improve the empirical performance of analytical CF models through trajectory-based parameter calibration \citep{Kesting2008} and the examination of heterogeneity in driving behavior \citep{Ossen2011}. Further research has highlighted the importance of incorporating human factors and additional behavioral mechanisms into existing formulations \citep{Saifuzzaman2014}. Despite these developments, the functional relationships underlying analytical CF models are generally specified through prior assumptions, with empirical observations primarily informing parameter estimation and model-specific refinements. Consequently, the behavioral responses and traffic flow dynamics represented by these models remain dependent on their prescribed mathematical structures. This structural dependence motivates the exploration of analytical CF relationships directly from empirical trajectory data, while retaining the interpretability and theoretical tractability of classical formulations.

\subsection{Data-Driven and Symbolic Discovery of CF Models}

Data-driven approaches have expanded the ability to represent complex and nonlinear CF behavior. For instance, \citet{Papathanasopoulou2015} employed locally weighted regression to estimate driving responses, while \citet{Huang2018} developed an LSTM-based CF model to capture asymmetric driving behavior and temporal dependencies. These approaches offer flexible representations of vehicle interactions, although their implicit functional relationships complicate the direct interpretation of behavioral mechanisms and the analytical derivation of traffic flow properties. Nevertheless, dedicated methods have been developed to examine their dynamical behavior, including the theoretical string stability framework for neural-network-based CF models proposed by \citet{Zhang2024NNStability}.

Data-driven equation discovery provides a complementary approach for identifying explicit mathematical relationships. The sparse identification of nonlinear dynamics (SINDy) framework demonstrates how governing equations can be recovered by selecting relevant terms from a candidate function library \citep{Brunton2016}. More flexible symbolic regression methods extend this principle to the exploration of mathematical structures. In CF modeling, \citet{SciNetCFM2025} combined neural representation learning with symbolic regression to identify analytical optimal-speed and desired-spacing relationships within established CF formulations, and examined their numerical performance and stability properties. More recently, \citet{NOVA2026} investigated recurring symbolic structures in large-scale freeway trajectory observations, including residual-guided extensions and the cross-site transfer of discovered features. These studies have advanced the identification of interpretable CF relationships and provided initial evidence of structural recurrence and transferability across traffic observations.

\subsection{LLM-Assisted and Agentic Scientific Discovery}

Recent advances in LLMs have introduced knowledge-guided approaches to scientific equation discovery. \citet{LLMSR2025} leveraged scientific priors encoded in LLMs to generate programmatic equation hypotheses, which were refined through evolutionary search and numerical parameter optimization. \citet{LaSR2024} introduced a learned concept library that extracts reusable mathematical concepts from evaluated expressions to guide subsequent symbolic exploration. In CF modeling, \citet{SRLLM2025} integrated retrieval-augmented domain knowledge with deep reinforcement learning to construct interpretable acceleration equations, reporting several candidate CF laws discovered from empirical NGSIM trajectories. These approaches demonstrate how scientific knowledge and accumulated search experience can facilitate the exploration of mathematical structures.

\begin{table*}[!hb]
\centering
\caption{Comparison of representative approaches to CF
modeling and empirical equation discovery.}
\label{tab:related_work}
\footnotesize
\setlength{\tabcolsep}{6pt}
\renewcommand{\arraystretch}{1.18}

\begin{tabularx}{\textwidth}{
@{}
>{\raggedright\arraybackslash}p{3.2cm}
>{\raggedright\arraybackslash}p{3.8cm}
>{\raggedright\arraybackslash}X
@{}}

\toprule
\textbf{Research direction} &
\textbf{Models/Methods} &
\textbf{Key characteristics}\\
\midrule

\multicolumn{3}{@{}l}{
\textit{\textbf{I. Analytical CF Models}}
}\\
\addlinespace[3pt]

Classical CF modeling &
GHR, Gipps, Newell, OVM, FVDM, IDM &
Explicit mathematical formulations based on
predefined behavioral assumptions and traffic flow theory\\

\addlinespace[4pt]

Trajectory-based calibration &
Genetic-algorithm calibration &
Parameter optimization within prescribed CF
structures using observed trajectories\\

\midrule

\multicolumn{3}{@{}l}{
\textit{\textbf{II. Data-Driven and Symbolic Discovery of CF Models}}
}\\
\addlinespace[3pt]

Data-driven CF modeling &
Locally weighted regression, LSTM &
Learning nonlinear vehicle-response relationships
from empirical trajectories\\

\addlinespace[4pt]

Neural-symbolic discovery &
SciNet-CFM &
Discovering analytical optimal-state relationships
through neural representations and symbolic regression\\

\addlinespace[4pt]

Automated symbolic discovery &
NOVA &
Symbolic structure search with structural recurrence,
residual-guided refinement, and cross-site transfer\\

\midrule

\multicolumn{3}{@{}l}{
\textit{\textbf{III. LLM-Assisted and Agentic Scientific Discovery}}
}\\
\addlinespace[3pt]

LLM-guided symbolic search &
LLM-SR, LaSR &
Using scientific priors and learned concepts
to guide mathematical expression search\\

\addlinespace[4pt]

LLM-assisted CF law discovery &
SR-LLM &
Retrieval-augmented domain knowledge and
reinforcement-learning-based symbolic discovery\\

\addlinespace[4pt]

Agentic model refinement &
TR-Agent &
Automated traffic model improvement through
hypothesis generation, implementation,
evaluation, and iterative refinement\\

\addlinespace[4pt]

Agentic equation discovery &
SR-Scientist, KeplerAgent &
Tool-driven equation refinement and
physics-guided symbolic exploration\\

\midrule
\multicolumn{3}{@{}l}{
\textit{\textbf{Our Study}}
}\\
\addlinespace[3pt]
\textbf{Knowledge-driven agentic CF law discovery} &
\textbf{AERM} &
\textbf{Classical CF knowledge-informed hypothesis generation, diagnosis-guided structural exploration, evolutionary search, and physics-aware selection}\\

\bottomrule

\end{tabularx}
\end{table*}

Recent research has further extended LLM-assisted equation discovery toward autonomous scientific reasoning and model refinement. \citet{TRAgent2025} developed an AI research agent for the iterative improvement of established traffic models, including IDM, MOBIL, and LWR, and evaluated the resulting refinements across independent real-world datasets. \citet{SRScientist2026} enabled autonomous data analysis, computational experimentation, and equation refinement through tool-based interaction and experimental feedback. Similarly, \citet{KeplerAgent2026} introduced physics-guided reasoning to infer intermediate physical properties and configure symbolic regression accordingly. Together, these studies establish the potential of knowledge-guided structural search, autonomous experimental feedback, and physically informed equation discovery. Building on these advances, the present study focuses on discovering a shared analytical CF structure from heterogeneous trajectory sources and examining its behavioral applicability and traffic flow dynamical properties.
Table~\ref{tab:related_work} summarizes representative approaches to CF modeling and empirical equation discovery.

% =========================================================
% 3. Methodology
% =========================================================
\section{Methodology}
\label{sec:methodology}

\begin{figure*}[!ht]
	\centering
	\includegraphics[width=\textwidth]{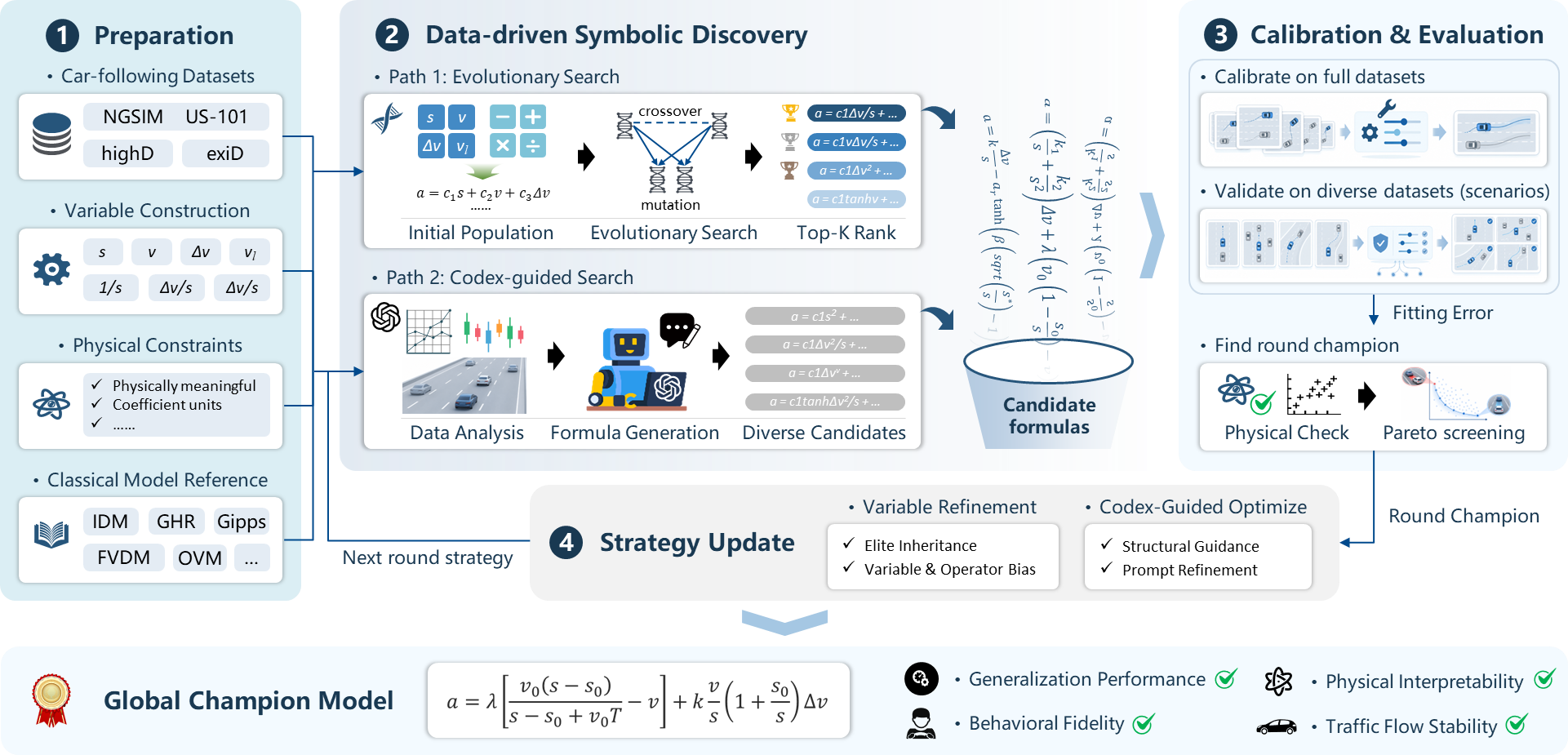}
	\caption{Overview of the proposed knowledge-driven agentic framework for CF law discovery.}
	\label{fig:framework}
\end{figure*}

To address the challenges of discovering calibration-promising and physically meaningful CF laws, this study develops a knowledge-driven agentic framework for empirical CF model discovery from real-world driving data. The framework integrates knowledge-driven hypothesis generation and state-dependent diagnostic feedback with complementary evolutionary symbolic search. Candidate equations are evaluated through two-stage parameter calibration, closed-loop trajectory reconstruction, physical consistency checks, and multi-objective Pareto selection. State-dependent reconstruction deficiencies are subsequently diagnosed using trajectory-specific analysis and fed back to the scientific-reasoning agent to guide iterative hypothesis refinement and structural exploration. The overall framework is illustrated in Fig.~\ref{fig:framework}.

% ---------------------------------------------------------
\subsection{Problem Formulation}
\label{sec:problem_formulation}
% ---------------------------------------------------------

Consider a CF system consisting of a follower vehicle $i$ and its preceding vehicle $i-1$. The longitudinal interaction is characterized by the follower speed $v_i(t)$, leader speed $v_{i-1}(t)$, net spacing $s_i(t)$, and relative speed $\Delta v_i(t)=v_{i-1}(t)-v_i(t)$, where $\Delta v_i(t)<0$ indicates that the follower is approaching the leader. Based on these quantities, the CF state is represented by:

\begin{equation}
	\mathbf{x}_i(t)=
	\left[
	v_i(t),
	v_{i-1}(t),
	\Delta v_i(t),
	s_i(t),
	h_i(t),
	\frac{1}{v_i^{+}(t)},
	\frac{1}{s_i^{+}(t)},
	\frac{\Delta v_i(t)}{s_i^{+}(t)}
	\right]^{\top},
	\label{eq:state_vector}
\end{equation}
where $h_i(t)=s_i(t)/v_i^{+}(t)$ denotes the time gap, with $v_i^{+}(t)=\max(v_i(t),\varepsilon_v)$ and $s_i^{+}(t)=\max(s_i(t),\varepsilon_s)$. The small positive constants $\varepsilon_v$ and $\varepsilon_s$ ensure numerical validity. These variables and their physically meaningful transformations constitute the candidate variable library for equation discovery.

The objective of the developed agentic framework is to autonomously discover an explicit analytical CF law that describes the longitudinal response of the follower vehicle:

\begin{equation}
    a_i(t)=
    F\left(
    \mathbf{x}_i(t-\tau);
    \boldsymbol{\theta}_{F,\tau}
    \right),
    \qquad
    F\in\mathcal{G}_B,\quad
    \tau\in\mathcal{T},\quad
    \tau=\ell\Delta t.
    \label{eq:discovery_problem}
\end{equation}
where $F$ denotes the unknown symbolic structure, $\boldsymbol{\theta}_{F,\tau}$ represents its adjustable parameters, and $\mathcal{G}_B$ defines the acceptable symbolic expression space. With a sampling interval of $\Delta t=0.1$ s, the candidate response delays are $\mathcal{T}=\{\ell\Delta t\mid\ell=1,\ldots,10\}$ \citep{Ma2006ReactionTime,Ranjitkar2011Response}. For each candidate delay, the observed CF states and follower accelerations are temporally aligned for parameter fitting and model evaluation. The training--validation split is performed at the original trajectory level before temporal alignment to prevent data leakage.

For each candidate formula-delay pair, preliminary parameter fitting is performed using the corresponding temporally aligned training observations:

\begin{equation}
    \boldsymbol{\theta}_{F,\tau}^{(0)}
    =
    \arg\min_{\boldsymbol{\theta}\in\Theta_F}
    \mathcal{L}_{\mathrm{acc}}
    \left(
    F,\tau,\boldsymbol{\theta};
    \mathcal{D}_{\mathrm{Train}}
    \right),
    \label{eq:parameter_calibration}
\end{equation}
where $\mathcal{L}_{\mathrm{acc}}$ denotes the acceleration-level fitting objective, and $\boldsymbol{\theta}_{F,\tau}^{(0)}$ represents the initial parameter estimates obtained during evolutionary search for each formula-delay pair.

For each candidate structure $F$ and response delay $\tau\in\mathcal{T}$, the corresponding follower dynamics are defined by:

\begin{equation}
    \begin{aligned}
        \dot{v}_i(t) &=
        F\left(
        \mathbf{x}_i(t-\tau);
        \boldsymbol{\theta}_{F,\tau}
        \right),\\
        \dot{s}_i(t) &=
        v_{i-1}(t)-v_i(t).
    \end{aligned}
    \label{eq:cf_dynamics}
\end{equation}

Given the observed leader trajectory and the required initial state history, the CF dynamics enable closed-loop reconstruction of the follower trajectory, providing a basis for evaluating candidate equations in terms of reconstruction accuracy and physical consistency.

\subsection{Knowledge-Driven Hypothesis Generation}
\label{sec:knowledge_driven_hypotheses}
To enable targeted structural refinement, the proposed framework employs a scientific-reasoning agent that draws on empirical driving evidence, domain knowledge, and historical search experience to formulate new behavioral hypotheses. By examining observed vehicle-response patterns and model deficiencies revealed through experimental evaluation, the agent identifies potential relationships and proposes directions for subsequent structural exploration.

\textbf{Traffic flow knowledge grounding.}
The agent is provided with foundational knowledge of traffic flow theory and CF dynamics, including the governing formulations and physical interpretations of classical CF models, namely GHR, Helly, Gipps, Newell, OVM, FVDM, and IDM. These formulations provide references for understanding established relationships among vehicle speed, spacing, relative motion, and acceleration, together with their underlying behavioral assumptions and parameter meanings. The agent also draws on fundamental concepts of traffic equilibrium and stability when formulating physically meaningful hypotheses. This knowledge provides a theoretical reference for interpreting empirical driving patterns and exploring new CF relationships, without prescribing a specific model structure.

\textbf{Historical structural memory.}
The framework maintains a record of competitive equations and their evaluation results across discovery rounds. These records inform subsequent hypothesis generation and provide candidate structures for evolutionary search.

\textbf{Empirical evidence integration.}
Empirical driving data and model evaluation results provide evidence for identifying limitations in candidate CF laws. Beyond overall reconstruction accuracy, the framework examines state-dependent error patterns and physical feasibility. Recurring discrepancies across different driving conditions and data sources are analyzed to identify behavioral relationships that may require further refinement.

At discovery round $r$, the agent combines traffic flow knowledge, historical search experience, and empirical evidence to formulate hypotheses for the next iteration:

\begin{equation}
    \mathcal{H}_{r+1}
    =
    \mathcal{R}_{\mathrm{agent}}
    \left(
    \mathcal{K},
    \mathcal{M}_r,
    \mathcal{E}_r
    \right),
    \label{eq:knowledge_driven_hypothesis}
\end{equation}
where $\mathcal{K}$ represents foundational traffic flow knowledge, $\mathcal{M}_r$ stores historical search information, and $\mathcal{E}_r$ contains empirical observations and model evaluation results. Based on these inputs, the agent formulates hypotheses about potential CF relationships and proposes corresponding symbolic expressions. The generated candidates are subsequently calibrated and evaluated, with the resulting evidence informing the next discovery round.

\subsection{Complementary Symbolic Exploration}
\label{sec:complementary_search}
To balance knowledge-driven structural refinement with diverse mathematical exploration, the proposed framework integrates two complementary search pathways. Knowledge-driven agentic search draws on classical CF knowledge and diagnosed model deficiencies to direct structural exploration, while independent evolutionary search explores alternative mathematical expressions. Both pathways complement each other, and all resulting candidates undergo a unified calibration and evaluation procedure.
 
Candidate CF laws are represented as symbolic expression trees, where leaf nodes consist of acceptable input variables and adjustable coefficients, and internal nodes represent mathematical operators. The operator library includes arithmetic operations and nonlinear functions:

\begin{equation}
    \begin{aligned}
        \mathcal{O}_{b} &=
        \{+,-,\times,\div,\mathrm{power}\},\\
        \mathcal{O}_{u} &=
        \{\exp,\log,\sqrt{\cdot},|\cdot|,\tanh\},
    \end{aligned}
    \label{eq:operator_library}
\end{equation}
where $\mathcal{O}_{b}$ and $\mathcal{O}_{u}$ denote the binary and unary operator sets, respectively. Together with the input variables defined in Section~\ref{sec:problem_formulation}, these operators define the acceptable expression space $\mathcal{G}_B$. Candidate expressions are subject to constraints on structural complexity, dimensional consistency, and numerical validity. Numerical coefficients are represented by symbolic placeholders during structure generation and subsequently determined through numerical optimization.

At each discovery round, candidate expressions are explored through two complementary pathways. The knowledge-driven agentic search translates the behavioral hypotheses generated in Section~\ref{sec:knowledge_driven_hypotheses} into mathematical expressions and further refines their structures through evolutionary optimization. In parallel, the evolutionary search generates and evolves candidate expressions within the acceptable symbolic space, incorporating competitive solutions retained from previous rounds. The resulting expressions are screened for structural validity, dimensional consistency, and numerical feasibility, and competitive candidates are retained for subsequent evaluation.

Let $\mathcal{S}_{r}^{p}$ denote the collection of initial populations associated with pathway $p$ at discovery round $r$, where $p\in\{A,E\}$ represents knowledge-driven agentic search and evolutionary search, respectively. The candidate pool is constructed as:

\begin{equation}
    \begin{aligned}
        \mathcal{C}_{r}^{p}
        &=
        \bigcup_{\mathcal{P}\in\mathcal{S}_{r}^{p}}
        \operatorname{Panel}_{K}
        \left[
        \operatorname{Evolve}_{\mathcal{G}_B}
        \left(
        \mathcal{P};
        \mathcal{D}_{\mathrm{Train}}
        \right)
        \right],
        \quad p\in\{A,E\},\\
        \mathcal{C}_{r}
        &=
        \mathcal{C}_{r}^{A}
        \cup
        \mathcal{C}_{r}^{E},
    \end{aligned}
    \label{eq:candidate_pool}
\end{equation}
where $\operatorname{Evolve}_{\mathcal{G}_B}$ represents evolutionary optimization within the acceptable symbolic space, and $\operatorname{Panel}_{K}$ selects up to $K$ competitive expressions from each evaluated population. The first discovery round is initialized through evolutionary search, while knowledge-driven agentic search is introduced from the second round after the initial experimental evidence becomes available.

The resulting candidate pool is subjected to parameter calibration, closed-loop trajectory reconstruction, and physics-aware evaluation, as described in Section~\ref{sec:calibration_eval}. Evaluation results are subsequently incorporated into the structural memory and experimental evidence for the next discovery round.

\subsection{Candidate Calibration and Physics-Aware Evaluation}
\label{sec:calibration_eval}

Candidate equations from both search pathways undergo two-stage parameter calibration. Numerical coefficients are initially fitted to observed accelerations and subsequently refined through bounded multi-start optimization of the cross-source closed-loop reconstruction objective defined in Eq.~\eqref{eq:parameter_fit}. The resulting parameter vector $\boldsymbol{\theta}_{F,\tau}^{G}$ is evaluated through recursive trajectory reconstruction using the observed leader motion and the required initial follower-state history. The discrete follower dynamics are updated as:

\begin{equation}
    \begin{aligned}
        \tilde a_{i,k}
        &=
        \operatorname{clip}
        \left(
        F(\hat{\mathbf{x}}_{i,k-\ell};
        \boldsymbol{\theta}_{F,\tau}^{G}),
        a_{\mathrm{clip}}^{\min},a_{\mathrm{clip}}^{\max}
        \right),\\
        \hat v_{i,k+1}
        &=
        \max\left(
        0,\hat v_{i,k}+\tilde a_{i,k}\Delta t
        \right),\\
        \hat s_{i,k+1}^{\mathrm{raw}}
        &=
        \hat s_{i,k}
        +
        \left(
        v_{i-1,k}-\hat v_{i,k}
        \right)\Delta t,\\
        \hat s_{i,k+1}
        &=
        \max\left(
        \hat s_{i,k+1}^{\mathrm{raw}},
        \varepsilon_s
        \right).
    \end{aligned}
    \label{eq:closed_loop_reconstruction}
\end{equation}
where $i$ and $i-1$ index the follower and leader vehicles, respectively, and $k$ denotes the discrete time step. The integer lag is $\ell=\tau/\Delta t$, with $\Delta t=0.1$ s. The delayed state $\hat{\mathbf{x}}_{i,k-\ell}$ is constructed from the reconstructed follower state and the observed leader trajectory, and the superscript $G$ denotes the globally calibrated parameter vector. A hat indicates a reconstructed quantity, while $\tilde a_{i,k}$ is the acceleration after clipping to $[a_{\mathrm{clip}}^{\min},a_{\mathrm{clip}}^{\max}]=[-8,4]$ m/s$^2$. The raw spacing update $\hat{s}_{i,k+1}^{\mathrm{raw}}$ is safeguarded by the minimum spacing $\varepsilon_s=0.5$ m.

For data source $d$ and partition $S\in\{\mathrm{Train},\mathrm{Val}\}$, the spacing reconstruction error is defined as:

\begin{equation}
    R_{s,d}^{S}(F,\tau,\boldsymbol{\theta})
    =
    \sqrt{
        \frac{
            \displaystyle
            \sum_{e\in\mathcal{D}_{d,S}}
            \sum_{k\in\mathcal{K}_e}
            (\hat{s}_{e,k}-s_{e,k})^2
        }{
            \displaystyle
            \sum_{e\in\mathcal{D}_{d,S}} |\mathcal{K}_e|
        }
    },
    \label{eq:source_spacing_rmse}
\end{equation}
where $\mathcal{D}_{d,S}$ contains the CF episodes from data source $d$ and partition $S$, with $e$ and $k$ indexing episodes and evaluated frames, respectively. The set $\mathcal{K}_e$ contains the evaluated frame indices of episode $e$, and $\hat{s}_{e,k}$ and $s_{e,k}$ denote the reconstructed and observed spacing, respectively. The reconstruction loss incorporates collision rate:

\begin{equation}
    L_d^S(F,\tau,\boldsymbol{\theta})
    =
    R_{s,d}^{S}(F,\tau,\boldsymbol{\theta})
    \left[
    1+c_d^S(F,\tau,\boldsymbol{\theta})
    \right],
    \label{eq:dataset_loss}
\end{equation}
where $c_d^S$ denotes the proportion of episodes containing at least one collision. Accordingly, $L_d^S$ represents the collision-penalized reconstruction loss for data source $d$. The cross-source calibration objective is:

\begin{equation}
    \begin{aligned}
        J^S(F,\tau,\boldsymbol{\theta})
        &=
        \frac{1}{D}
        \sum_{d=1}^{D}
        L_d^S(F,\tau,\boldsymbol{\theta}),\\
        \boldsymbol{\theta}_{F,\tau}^{G}
        &=
        \arg\min_{\boldsymbol{\theta}\in\Theta_F}
        J^{\mathrm{Train}}(F,\tau,\boldsymbol{\theta}).
    \end{aligned}
    \label{eq:parameter_fit}
\end{equation}
where $D$ is the number of discovery data sources, and $J^S$ denotes the average reconstruction loss across these sources. The set $\Theta_F$ defines the admissible parameter bounds for formula $F$. The globally calibrated parameter vector $\boldsymbol{\theta}_{F,\tau}^{G}$ is obtained by minimizing the training objective and remains fixed during validation evaluation.

Candidate equations are subject to six physical consistency criteria, including effective CF-variable dependence, dimensional consistency, causal inputs, numerical validity, directional response consistency, and valid closed-loop simulation.

\begin{equation}
    \mathcal{Z}_r=\mathcal{C}_r\times\mathcal{T},
    \label{eq:formula_delay_space}
\end{equation}
where $r$ denotes the discovery round, $\mathcal{C}_r$ is the set of candidate formula structures generated in round $r$, and $\mathcal{T}$ is the set of candidate response delays. Their Cartesian product $\mathcal{Z}_r$ contains all candidate formula--delay pairs considered in that round. The physically feasible subset is defined as:

\begin{equation}
    \mathcal{F}_r
    =
    \left\{
    (F,\tau)\in\mathcal{Z}_r:
    \prod_{h\in\mathcal{I}_{\mathrm{adm}}}
    H_h(F,\tau)=1,\quad
    \max_d c_d^{\mathrm{Val}}
    (F,\tau,\boldsymbol{\theta}_{F,\tau}^{G})
    \leq c_{\max}
    \right\},
    \label{eq:physical_feasibility}
\end{equation}
where $\mathcal{I}_{\mathrm{adm}}$ indexes the six physical consistency checks, and $H_h(F,\tau)\in\{0,1\}$ indicates whether the $h$-th check is satisfied. The threshold $c_{\max}=0.05$ specifies the maximum permitted validation collision rate across discovery sources. Thus, $\mathcal{F}_r$ contains candidate formula--delay pairs that pass all six checks and satisfy this collision-rate constraint.

Although physical admissibility checks establish basic requirements for candidate equations, feasible candidates may still differ in their representation of CF mechanisms and interpretability. To account for these differences, physical interpretability is evaluated through four dimensions: numerical behavior, CF mechanism-related feature coverage, directional response consistency, and interpretation evidence. The overall interpretability score is defined as:

\begin{equation}
    \begin{aligned}
    S_{\mathrm{int}}(F)
    ={}&
    S_{\mathrm{num}}(F)
    +S_{\mathrm{mech}}(F)\\
    &+S_{\mathrm{dir}}(F)
    +S_{\mathrm{evid}}(F),
    \end{aligned}
    \qquad
    0\leq S_{\mathrm{int}}(F)\leq100.
    \label{eq:interpretability_score}
\end{equation}

\begin{equation}
    \begin{aligned}
    S_{\mathrm{num}}(F)
    &=
    10I_{\mathrm{finite}}
    +4I_{\mathrm{spread}}
    +4I_{\mathrm{bounded}},\\[3pt]
    S_{\mathrm{mech}}(F)
    &=
    \min\left(
    22,\,
    9I_{\mathrm{closing}}
    +5I_{\mathrm{relative}}
    +6I_{\mathrm{spacing}}
    +2I_{\mathrm{free}}
    \right),\\[3pt]
    S_{\mathrm{dir}}(F)
    &=
    13I_1+13I_2+4I_3,\\[3pt]
    S_{\mathrm{evid}}(F)
    &=4I_{\mathrm{summary}}
    +\min(6,2N_{\mathrm{term}})\\
    &\quad+\min(12,4N_{\mathrm{model}})
    +8I_{\mathrm{caveat}}.
    \end{aligned}
    \label{eq:interpretability_components}
\end{equation}

The four interpretability components contribute up to 18, 22, 30, and 30 points, respectively. Each indicator $I_{\cdot}$ is binary, taking a value of one when its corresponding criterion is satisfied and zero otherwise. Numerical behavior is evaluated over 27 combinations of follower speed $v\in\{5,20,35\}$ m/s, spacing $s\in\{5,15,40\}$ m, and relative speed $\Delta v\in\{-4,0,4\}$ m/s. $I_{\mathrm{finite}}$ requires all raw acceleration outputs, evaluated before clipping, to be finite. $I_{\mathrm{spread}}$ additionally requires the difference between the maximum and minimum outputs to exceed $10^{-4}$ m/s$^2$, while $I_{\mathrm{bounded}}$ requires their maximum absolute value to be no greater than $8$ m/s$^2$. The latter two indicators are evaluated only when all outputs are finite.

Mechanism coverage is determined by the presence of predefined atomic features in the symbolic expression tree. Specifically, $I_{\mathrm{closing}}$ checks for the $\Delta v/s$ feature, $I_{\mathrm{relative}}$ for relative speed or leader speed, $I_{\mathrm{spacing}}$ for spacing, inverse spacing, or time headway, and $I_{\mathrm{free}}$ for follower speed or its inverse. Each indicator is activated only by its corresponding predefined feature; composite features are not decomposed into their constituent variables. Directional consistency is evaluated through three fixed comparisons:

\begin{equation*}
\begin{aligned}
I_1 &= \mathbb{I}\left[
F(20,20,-5)\leq F(20,20,-1)+0.05
\right],\\
I_2 &= \mathbb{I}\left[
F(20,8,-3)\leq F(20,35,-3)+0.05
\right],\\
I_3 &= \mathbb{I}\left[
F(20,20,3)\geq F(20,20,-3)-0.05
\right].
\end{aligned}
\end{equation*}
where $F(v,s,\Delta v)$ denotes the acceleration evaluated with fixed candidate coefficients, and $\mathbb{I}[\cdot]$ denotes a binary indicator function. The tolerance is $0.05$ m/s$^2$. The three comparisons examine responses to stronger closing motion, reduced spacing, and separating motion, respectively.

For interpretation evidence, Agent generates formula summaries, term-level explanations, links to classical CF models, and limitation statements. $I_{\mathrm{summary}}$ and $I_{\mathrm{caveat}}$ indicate the presence of a nonempty summary and at least one limitation statement, respectively. $N_{\mathrm{term}}$ counts explanations validated against the formula structure and associated physical mechanisms, while $N_{\mathrm{model}}$ counts distinct, structurally validated links to classical CF models. Each count contributes up to three items, as specified by the caps in Eq.~\eqref{eq:interpretability_components}.

Feasible candidates are evaluated using four objectives that account for average cross-source reconstruction loss, worst-source reconstruction loss, collision rate, and physical interpretability. Pareto dominance is employed to identify non-dominated candidates, from which the global champion is selected using a lexicographic minimax-regret strategy. The four objectives for each formula--delay pair $z=(F,\tau)$ are defined as:

\begin{equation}
    \mathbf{q}(F,\tau)=
    \left(
    \begin{array}{c}
    J^{\mathrm{Val}}(F,\tau,\boldsymbol{\theta}_{F,\tau}^{G})\\
    \displaystyle\max_d L_d^{\mathrm{Val}}
    (F,\tau,\boldsymbol{\theta}_{F,\tau}^{G})\\
    \displaystyle\max_d c_d^{\mathrm{Val}}
    (F,\tau,\boldsymbol{\theta}_{F,\tau}^{G})\\
    100-S_{\mathrm{int}}(F)
    \end{array}
    \right),
    \label{eq:robust_objectives}
\end{equation}
where all four objectives are minimized, with the interpretability objective expressed as $100-S_{\mathrm{int}}(F)$ to favor physically interpretable candidates. The Pareto front is defined as:

\begin{equation}
    \mathcal{P}_r
    =
    \left\{
    z\in\mathcal{F}_r:
    \nexists z'\in\mathcal{F}_r
    \text{ such that }
    \mathbf{q}(z')\prec\mathbf{q}(z)
    \right\},
    \label{eq:pareto_front}
\end{equation}
where $\prec$ denotes Pareto dominance. The global champion at discovery round $r$ is selected from the Pareto front using a lexicographic minimax-regret strategy:

\begin{equation}
\begin{aligned}
z_r^*
&=
\underset{z_i\in\mathcal{P}_r}
{\operatorname{arg\,min}_{\mathrm{lex}}}
\left(R_i^{\max},R_i^{\mathrm{mean}}\right),\\
R_i^{\max}
&=\max_{1\leq j\leq4}r_{ij},\\
R_i^{\mathrm{mean}}
&=\frac{1}{4}\sum_{j=1}^{4}r_{ij},
\end{aligned}
\label{eq:round_champion_selection}
\end{equation}
where $z_r^*=(F_r^*,\tau_r^*)$ denotes the selected formula--delay pair at round $r$, $z_i\in\mathcal{P}_r$ indexes the Pareto candidates, and $j$ indexes the four objectives. $R_i^{\max}$ and $R_i^{\mathrm{mean}}$ denote the maximum and mean normalized regrets, respectively.

The normalized regret of candidate $z_i$ on objective $j$ is calculated as:

\begin{equation}
    r_{ij}
    =
    \frac{f_{ij}-f_j^{\min}}
    {\max\left(
    10^{-12},
    f_j^{\max}-f_j^{\min}
    \right)},
    \label{eq:normalized_regret}
\end{equation}
where $f_{ij}=q_j(z_i)$ denotes the $j$-th objective value of candidate $z_i$. The quantities $f_j^{\min}$ and $f_j^{\max}$ are the minimum and maximum values of objective $j$ among the Pareto candidates. The normalized regret $r_{ij}$ measures the relative deviation from the best value of that objective. The constant $10^{-12}$ prevents division by zero or an excessively small denominator. The lexicographic ordering first minimizes the maximum normalized regret and then uses the mean normalized regret as a secondary criterion.

\subsection{Diagnostic Feedback and Model Selection}
\label{sec:diagnostic_feedback_selection}
At the end of each discovery round, the global champion $z_r^*=(F_r^*,\tau_r^*)$ is selected through four-objective Pareto evaluation. Its parameters are then independently recalibrated for each trajectory in a fixed diagnostic panel, while retaining the selected formula structure. The resulting reconstructions are analyzed to identify driving conditions associated with systematic reconstruction errors.

For each diagnostic window $w$, the champion model is compared with classical reference models, as the best-performing model in terms of spacing reconstruction error selected as the reference:

\begin{equation}
    \begin{aligned}
        m_w^*
        &=
        \arg\min_{m\in\mathcal{M}_{\mathrm{ref}}}
        \frac{1}{|I_w|}
        \sum_{k\in I_w}
        (\hat{s}_{m,k}-s_k)^2,\\
        y_w
        &=
        \frac{1}{|I_w|}
        \sum_{k\in I_w}
        \left[
        (\hat{s}_{F_r^*,\tau_r^*,k}-s_k)^2
        -
        (\hat{s}_{m_w^*,k}-s_k)^2
        \right],
    \end{aligned}
    \label{eq:regional_target}
\end{equation}
where $\hat{s}_{F_r^*,\tau_r^*,k}$ and $\hat{s}_{m_w^*,k}$ denote the reconstructed spacings of the champion and the best-performing reference model within window $w$, respectively. $\mathcal{M}_{\mathrm{ref}}$ denotes the set of classical reference models, and $I_w$ contains the evaluated frames. Positive values indicate driving conditions under which the champion exhibits larger reconstruction errors than the best-performing reference model.

A shallow decision tree relates the window-level error differences to vehicle spacing, speed, relative motion, and recent state changes.  Recurring reconstruction-error patterns across vehicles and data sources, together with physical evaluation results, are used to diagnose systematic model deficiencies and guide the agent in formulating subsequent structural hypotheses. In parallel, evolutionary search continues to explore alternative expressions. Candidates from both pathways undergo the same calibration, physical screening, and Pareto selection procedure, while five competitive and structurally distinct equations are retained in the historical memory.

Search convergence is monitored separately through the development reconstruction loss of each round's champion. The lowest loss recorded across successive rounds is tracked to determine whether further improvement is achieved. Let

\begin{equation}
    \begin{aligned}
        E_r &=
        J^{\mathrm{Val}}
        \left(
        F_r^*,\tau_r^*,
        \boldsymbol{\theta}_{F_r^*,\tau_r^*}^{G}
        \right),\\
        E_r^* &=
        \min_{1\leq j\leq r} E_j,
    \end{aligned}
    \label{eq:historical_best_loss}
\end{equation}
where $E_r$ denotes the development reconstruction loss of the selected formula-delay pair in round $r$, and $E_r^*$ represents the lowest development reconstruction loss recorded across all selected pairs up to round $r$. The relative improvement \(\eta_r\) therefore measures the reduction in the best-so-far reconstruction loss between consecutive discovery rounds. The relative improvement is defined as:

\begin{equation}
    \eta_r =
    \frac{E_{r-1}^*-E_r^*}{E_{r-1}^*},
    \qquad r\geq 2.
    \label{eq:search_improvement}
\end{equation}

The discovery process terminates when the relative
improvement remains below $\epsilon_{\mathrm{stop}}=0.02$ for two consecutive rounds. Runs reaching the maximum search budget before satisfying this criterion are recorded as nonconverged.
% \begin{equation}
%     \eta_{r-1}<\epsilon_{\mathrm{stop}},
%     {and} 
%     \eta_r<\epsilon_{\mathrm{stop}},
%     \label{eq:search_stopping}
% \end{equation}
Upon termination at round $R$, the final discovered CF law is given by the global champion $z^*=z_R^*=(F_R^*,\tau_R^*)$. Its mathematical structure and response delay are retained for external evaluation, while the numerical coefficients are recalibrated according to the protocols described in Section~\ref{sec:baselines}.

% =========================================================
% 4. Experimental design
% =========================================================
\section{Experimental Design}
\label{sec:experimental_design}

\subsection{Datasets and Experimental Setup}
\label{sec:datasets}
Six real-world trajectory datasets are used for empirical CF law discovery and external evaluation, where NGSIM US-101 \citep{FHWA2006}, highD \citep{Krajewski2018}, and exiD  \citep{Moers2022} are employed for equation discovery across different roadway and traffic conditions, while NDS \citep{Zhu2018ShanghaiNDS, zhang2021comprehensive}, NGSIM I-80 \citep{FHWA2006}, and Waymo trajectory data \citep{Ettinger2021Waymo, hu2022processing} are used for external evaluation. 

Vehicle trajectories are processed to extract valid and noise-free leader-follower episodes, from which spacing, speed, relative speed, and acceleration are obtained using consistent physical units and sampling conventions. For each discovery dataset (NGSIM US-101, highD, and exiD), 80\% of the valid CF episodes are assigned to the training (Train) partition, with the remaining 20\% assigned to the development (Val) partition. Train is used for equation discovery and parameter calibration, while Val is used for candidate evaluation and selection. To support state-dependent error diagnosis and subsequent hypothesis refinement, a fixed panel of 200 CF episodes per discovery dataset is used for trajectory-specific recalibration and error analysis. NDS, NGSIM I-80, and Waymo datasets are reserved for external evaluation of the discovered CF law.
The datasets and main experimental settings are summarized in Table~\ref{tab:experimental_setup}.

\begin{table}[!t]
    \centering
    \caption{Datasets and experimental configuration
    for empirical CF law discovery and evaluation.}
    \label{tab:experimental_setup}
    \small
    \setlength{\tabcolsep}{6pt}
    \renewcommand{\arraystretch}{1.12}

    \begin{tabularx}{\linewidth}{
        @{}
        >{\raggedright\arraybackslash}p{0.43\linewidth}
        >{\raggedright\arraybackslash}X
        @{}
    }
        \toprule
        \textbf{Dataset / Component} &
        \textbf{Description / Configuration}\\
        \midrule

        \multicolumn{2}{@{}l}{
            \textit{Discovery datasets}
        }\\
        \addlinespace[2pt]

        NGSIM US-101 & Freeway traffic\\
        highD & Freeway traffic\\
        exiD & Freeway entry and exit areas\\

        \addlinespace[3pt]
        \midrule

        \multicolumn{2}{@{}l}{
            \textit{External evaluation datasets}
        }\\
        \addlinespace[2pt]

        NDS & Naturalistic driving on Freeway\\
        NGSIM I-80 & Freeway traffic\\
        Waymo & Autonomous vehicle traffic\\

        \addlinespace[3pt]
        \midrule

        \multicolumn{2}{@{}l}{
            \textit{Discovery settings}
        }\\
        \addlinespace[2pt]
        
        Reasoning agent backbone & GPT-5.6\\
        Sampling interval & $0.1$ s\\
        Subtree crossover probability & $0.4$\\
        Coefficient optimization &
        Bounded multi-start Nelder-Mead\\
        Acceleration clipping bounds &
        $[-8,4]$ m/s$^2$\\
        Minimum safeguarded spacing & $0.5$ m\\
        Directional-test tolerance &
        $0.05$ m/s$^2$\\
        Trajectory-specific diagnostic panel &
        200 episodes per discovery source\\
        Convergence criterion &
        $<2\%$ improvement for two consecutive rounds\\
        Completed discovery rounds & $16$\\
        \bottomrule
    \end{tabularx}
\end{table}

\subsection{Baseline Models}
\label{sec:baselines}

The discovered empirical CF law is compared with seven classical analytical models, including GHR \citep{Gazis1961}, Helly \citep{Helly1959}, Gipps \citep{Gipps1981}, Newell \citep{Newell2002}, OVM \citep{Bando1995}, FVDM \citep{Jiang2001}, and IDM \citep{Treiber2000}. Four additional baselines are included from recent automated equation discovery and model refinement studies: three discovered CF equations from SR-LLM \citep{SRLLM2025} and the Improved IDM from TR-Agent \citep{TRAgent2025}. These models represent different behavioral assumptions and longitudinal response mechanisms, providing reference formulations for assessing the empirical performance of the discovered mathematical structure. The corresponding model formulations and parameter definitions are provided in the Appendix.

\subsection{Calibration and Evaluation}
The calibration process is following \cite{punzo2021calibration}, where all models are calibrated using a real-coded genetic algorithm (GA) with model-specific parameter bounds. The same optimization procedure is applied under both calibration protocols. The calibration objective combines normalized spacing, speed, and acceleration reconstruction errors:

\begin{equation}
    J_{\mathrm{fit}}
    =
    \mathrm{NRMSE}_{s}
    +
    \mathrm{NRMSE}_{v}
    +
    \mathrm{NRMSE}_{a},
    \label{eq:ga_calibration_objective}
\end{equation}
where the normalized spacing and speed errors are defined as:

\begin{equation}
    \mathrm{NRMSE}_{z}
    =
    \sqrt{
    \frac{
        \sum_{n=1}^{N}
        (\hat{z}_n-z_n)^2
    }{
        \sum_{n=1}^{N}z_n^2
    }
    },
    \qquad z\in\{s,v\},
    \label{eq:ga_normalized_rmse}
\end{equation}
and the normalized acceleration error is calculated as:

\begin{equation}
    \mathrm{NRMSE}_{a}
    =
    \sqrt{
    \frac{
        \sum_{n=1}^{N}
        (\hat{a}_n-a_n)^2
    }{
        \sum_{n=1}^{N}
        (a_n+a_{\mathrm{off}})^2
    }
    },
    \label{eq:ga_normalized_acceleration_rmse}
\end{equation}
where $N$ denotes the total number of evaluated observations, and the hatted and unhatted variables represent reconstructed and observed quantities, respectively. The acceleration normalization uses $a_{\mathrm{off}}=9.81$ m/s$^2$ to avoid a near-zero reference-energy denominator. This offset affects only the normalization and does not alter the acceleration residuals. Collision events and invalid simulations are penalized during optimization. The GA employs tournament selection, elitist retention, crossover, and mutation. The main optimization settings are summarized in Table~\ref{tab:ga_settings}.

\begin{table}[!t]
    \centering
    \caption{GA configuration for external CF model calibration.}
    \label{tab:ga_settings}
    \small
    \setlength{\tabcolsep}{12pt}   
    \renewcommand{\arraystretch}{1.12}
    \begin{tabularx}{\linewidth}{@{}l@{\hspace{2em}}X@{}}
        \toprule
        \textbf{Parameter} & \textbf{Configuration}\\
        \midrule
        Encoding & Real-valued\\
        Population size & 64\\
        Maximum generations & 10,000\\
        Elite individuals & 4\\
        Selection & Tournament (size 3)\\
        Crossover probability & 0.9\\
        Mutation probability & 0.2\\
        Early stopping & 20 generations\\
        Improvement threshold & $10^{-4}$\\
        Random seed & 42\\
        \bottomrule
    \end{tabularx}
\end{table}

\textbf{Dataset-level calibration.}
For each model, a single parameter vector is calibrated using the CF trajectories within each evaluation dataset \citep{sun2020relationship}. The calibrated parameters are held fixed across all trajectories in that dataset. This protocol evaluates the ability of each model to reproduce heterogeneous driving behaviors using dataset-level shared parameters.

\textbf{Trajectory-specific calibration.}
Each model is independently calibrated for every CF trajectory and subsequently used to reconstruct that
trajectory using the observed leader motion and the required initial follower-state history \citep{Punzo2012, Treiber2013}. This protocol evaluates the fitting capability of each analytical structure under trajectory-specific parameter optimization.

% \subsection{Evaluation Metrics and Validation Protocols}
% \label{sec:evaluation_protocols}

The performance of the discovered CF law is evaluated through closed-loop trajectory reconstruction, using spacing RMSE, speed RMSE, and collision rate. Given the recorded leader trajectory and the required initial follower-state history, the follower trajectory is simulated recursively without resetting its state to subsequent observations. 
For $N$ evaluated observations, spacing and speed RMSE are defined as:

\begin{equation}
    \mathrm{RMSE}_{s}
    =
    \sqrt{
        \frac{1}{N}
        \sum_{n=1}^{N}
        (\hat{s}_n-s_n)^2
    },
    \label{eq:gap_rmse}
\end{equation}

\begin{equation}
    \mathrm{RMSE}_{v}
    =
    \sqrt{
        \frac{1}{N}
        \sum_{n=1}^{N}
        (\hat{v}_n-v_n)^2
    },
    \label{eq:speed_rmse}
\end{equation}
where $\hat{s}_n$ and $\hat{v}_n$ denote the reconstructed spacing and speed, respectively, and $s_n$ and $v_n$ are the corresponding observations. Collision rate is measured as the proportion of reconstructed episodes containing at least one contact event.

Beyond trajectory reconstruction, the physical properties of AERM are examined through equilibrium analysis, local and string stability analysis under the zero-delay assumption, and numerical simulations of platoon disturbance propagation. The corresponding results are presented in Section~\ref{sec:aerm_physical_analysis}.

\section{Automated Discovery of AERM}
\label{sec:results}
\subsection{Formulation of AERM}
\label{sec:aerm_discovery}
The proposed agentic framework identifies an explicit analytical CF law, termed the Adaptive Equilibrium Response Model (AERM), through iterative symbolic exploration and knowledge-driven structural refinement. The discovered acceleration equation is:

\begin{equation}
    a =
    \lambda
    \left[
    \frac{v_0(s-s_0)}
    {s-s_0+v_0T}
    -v
    \right]
    +
    k\frac{v}{s}
    \left(1+\frac{s_0}{s}\right)\Delta v,
    \label{eq:aerm_model}
\end{equation}
where $a$, $s$, $v$, and $\Delta v$ denote acceleration, net spacing, follower speed, and relative speed, respectively, with $\Delta v=v_{\mathrm{lead}}-v$. The five model parameters and their physical interpretations are summarized in
Table~\ref{tab:aerm_parameters}. The response delay selected during equation discovery is $\tau^*=0.1$ s, which is applied to the state inputs during delayed trajectory reconstruction.

\begin{table}[!t]
	\centering
	\caption{Parameters and physical interpretations of AERM.}
	\label{tab:aerm_parameters}
	\small
	\begin{tabular*}{\linewidth}{@{\extracolsep{\fill}} l l l l @{}}
		\toprule
		\textbf{Parameter} & \textbf{Unit} & \textbf{Physical interpretation} & \textbf{Calibration range}\\
		\midrule
		\(k\) & -- & Relative-speed response intensity & \([0, 5.0]\) \\
		\(\lambda\) & s\(^{-1}\) & Equilibrium-speed restoration rate & \([0, 1.0]\) \\
		\(v_0\) & m/s & Desired free-flow speed & \([5.0, 40.0]\) \\
		\(s_0\) & m & Standstill spacing & \([0.1, 15.0]\) \\
		\(T\) & s & Desired time gap & \([0.1, 5.0]\) \\
		\bottomrule
	\end{tabular*}
\end{table}

AERM combines equilibrium-speed restoration with state-dependent relative-speed feedback. Its first term describes the relaxation of follower speed toward a spacing-dependent equilibrium speed:

\begin{equation}
    V_e(s)=
    \frac{v_0(s-s_0)}
    {s-s_0+v_0T}.
\end{equation}

This function satisfies $V_e(s_0)=0$ and approaches $v_0$ as spacing increases. Near the standstill spacing, it can be approximated by the linear relationship $V_e(s)\approx(s-s_0)/T$, indicating that $T$ determines the local slope of the equilibrium speed--spacing relationship \citep{TreiberKesting2025}. The parameter $\lambda$ controls the rate of adjustment toward the equilibrium speed, with $1/\lambda$ representing the characteristic relaxation time of the equilibrium-speed restoration term. The second term describes the response to relative motion through a state-dependent gain:

\begin{equation}
    G(v,s)=
    k\frac{v}{s}\left(1+\frac{s_0}{s}\right).
\end{equation}

The coefficient $k$ determines the overall scale of relative-speed sensitivity. For positive parameter values, the gain increases with follower speed and decreases with net spacing, holding the other state variable fixed. Consequently, relative-speed feedback becomes more pronounced at higher speeds and shorter following distances. The sign of $\Delta v$ determines the direction of this response, as negative relative speed produces a deceleration contribution, whereas positive relative speed produces an acceleration contribution.

\subsection{AERM as a Member of the Optimal-Velocity Family}
\label{sec:aerm_ovm_idm}

According to Eq.~\eqref{eq:aerm_model}, the discovered AERM belongs to the family of optimal‑velocity‑type CF models represented by OVM and FVDM. Models within this family follow two core behavioral mechanisms, i.e., relaxation of follower velocity toward a spacing‑dependent equilibrium speed, and additional feedback responding to relative speed between leader and follower. In AERM, the first term performs equilibrium speed relaxation:

\begin{equation}
a=\lambda\left[V_e(s)-v\right]+k \frac{v}{s}\left(1+\frac{s_0}{s}\right) \Delta v
\end{equation}
where \(V_e(s)=\frac{v_{0}\left(s-s_{0}\right)}{s-s_{0}+v_{0} T}\) denotes the analytical equilibrium speed dictated by model parameters. The second term implements state‑dependent relative‑speed feedback. Despite sharing this high‑level structure, critical structural and behavioral differences distinguish AERM from OVM and FVDM.

Both the classical OVM framework and AERM establish
an explicit relationship between microscopic CF dynamics
and macroscopic equilibrium characteristics. A distinctive feature of AERM is that its equilibrium-speed function adopts a rational form parameterized by physically interpretable microscopic CF parameters. The rational‑form equilibrium speed naturally saturates toward free‑flow speed at large spacing and drops to zero at standstill spacing \(s_0\), requiring no manually tuned saturation or switching terms. The closed‑form flow‑density relationship can be directly computed without numerical integration. A trade-off accompanies this formulation, as the prescribed rational form of the equilibrium-speed function limits the flexibility of the resulting fundamental diagram. For traffic sites exhibiting strongly non‑rational empirical fundamental‑diagram shapes, pure parameter tuning may not fully reproduce observed macroscopic patterns.

FVDM extends OVM by adding a constant‑gain relative‑speed feedback term, which supplies fixed damping regardless of CF conditions. Under FVDM, a vehicle responds to a given relative speed with identical magnitude whether the leading vehicle is far ahead or very close. AERM instead makes relative-speed sensitivity state-dependent, with response strength increasing with follower speed and decreasing with increasing net spacing. This aligns with intuitive driver behavior, as anticipatory responses become stronger under tight following conditions and weaken for large time gaps. This state dependence also reshapes stability properties. The damping contribution from the relative‑speed term diminishes when spacing becomes large, and AERM asymptotically converges toward pure OVM dynamics under long‑gap conditions. Mathematically, existing models can be recovered as special nested cases of AERM, when setting the response coefficient \(k=0\) reduces AERM exactly to an OVM‑style formulation.

Beyond comparison with OVM and FVDM, AERM differs from many classical models in how its macroscopic properties emerge. Rather than postulating macroscopic relationships first, equilibrium speed, critical density, jam density and capacity are all derived from the microscopic interaction rules discovered from multi‑source trajectory data.

\subsection{Traffic Dynamic Properties of AERM}
\label{sec:aerm_physical_analysis}
A physically meaningful CF law should exhibit plausible equilibrium relationships and interpretable traffic dynamic properties across different temporal and spatial scales. Accordingly, this section evaluates the macroscopic equilibrium characteristics and microscopic dynamic responses of AERM, together with its local and string stability and platoon-level disturbance propagation properties.

\subsubsection{Macroscopic Equilibrium Analysis}
\label{sec:aerm_macroscopic}

Under homogeneous equilibrium conditions, all vehicles travel at the same speed with constant spacing, satisfying:

\begin{equation}
    \Delta v^{*}=0,\qquad a^{*}=0.
    \label{eq:aerm_equilibrium_condition}
\end{equation}

For $\lambda>0$, substituting these conditions into Eq.~\eqref{eq:aerm_model} yields the equilibrium speed-spacing relationship:

\begin{equation}
    v^{*}=V_e(s^{*})
    =
    \frac{v_0(s^{*}-s_0)}
    {s^{*}-s_0+v_0T}.
    \label{eq:aerm_equilibrium_relation}
\end{equation}

For $s^{*}\geq s_0$ and positive model parameters, the equilibrium speed increases monotonically with spacing and approaches the desired free-flow speed $v_0$ as spacing increases. At $s^{*}=s_0$, the equilibrium speed becomes zero, corresponding to the standstill equilibrium state.

For a homogeneous traffic stream with vehicle length $L_{\mathrm{veh}}$, the relationship between net spacing and traffic density is

\begin{equation}
    s^{*}=
    \frac{1}{\rho}-L_{\mathrm{veh}},
    \label{eq:aerm_spacing_density}
\end{equation}
where $\rho$ denotes traffic density and $L_{\mathrm{veh}}$ is the vehicle length. Substituting Eq.~\eqref{eq:aerm_spacing_density} into Eq.~\eqref{eq:aerm_equilibrium_relation} gives the equilibrium speed-density relationship:

\begin{equation}
    V_e(\rho)=
    v_0
    \frac{
        1-(L_{\mathrm{veh}}+s_0)\rho
    }{
        1+(v_0T-L_{\mathrm{veh}}-s_0)\rho
    }.
    \label{eq:aerm_speed_density}
\end{equation}

The corresponding equilibrium flow-density relationship is:

\begin{equation}
    q_e(\rho)=
    v_0\rho
    \frac{
        1-(L_{\mathrm{veh}}+s_0)\rho
    }{
        1+(v_0T-L_{\mathrm{veh}}-s_0)\rho
    }.
    \label{eq:aerm_flow_density}
\end{equation}

The jam density and critical density associated with the maximum equilibrium flow are respectively given by

\begin{equation}
    \rho_j=
    \frac{1}{L_{\mathrm{veh}}+s_0},
    \label{eq:aerm_jam_density}
\end{equation}

\begin{equation}
    \rho_c=
    \frac{1}{
        L_{\mathrm{veh}}+s_0+
        \sqrt{(L_{\mathrm{veh}}+s_0)v_0T}
    }.
    \label{eq:aerm_critical_density}
\end{equation}

By introducing a dimensionless parameter group
\begin{equation}
    \chi=\frac{v_0T}{L_{\mathrm{veh}}+s_0},
    \label{eq:aerm_dimensionless_group}
\end{equation}
the critical density, critical speed, and maximum
equilibrium flow can be expressed in normalized form as
\begin{equation}
    \begin{aligned}
        \frac{\rho_c}{\rho_j}
        &=\frac{1}{1+\sqrt{\chi}},\\
        \frac{v_c}{v_0}
        &=\frac{1}{1+\sqrt{\chi}},\\
        \frac{q_{\max}}{v_0\rho_j}
        &=\frac{1}{(1+\sqrt{\chi})^2},
    \end{aligned}
    \label{eq:aerm_dimensionless_critical}
\end{equation}
where $v_c=V_e(\rho_c)$ denotes the critical speed and $q_{\max}=\rho_c v_c$ denotes the maximum equilibrium flow. These relationships reveal that the normalized critical traffic characteristics of AERM are governed by a single dimensionless group, linking the microscopic car-following parameters to macroscopic traffic flow properties.

\begin{figure*}[!t]
    \centering
    \includegraphics[width=0.48\textwidth]
    {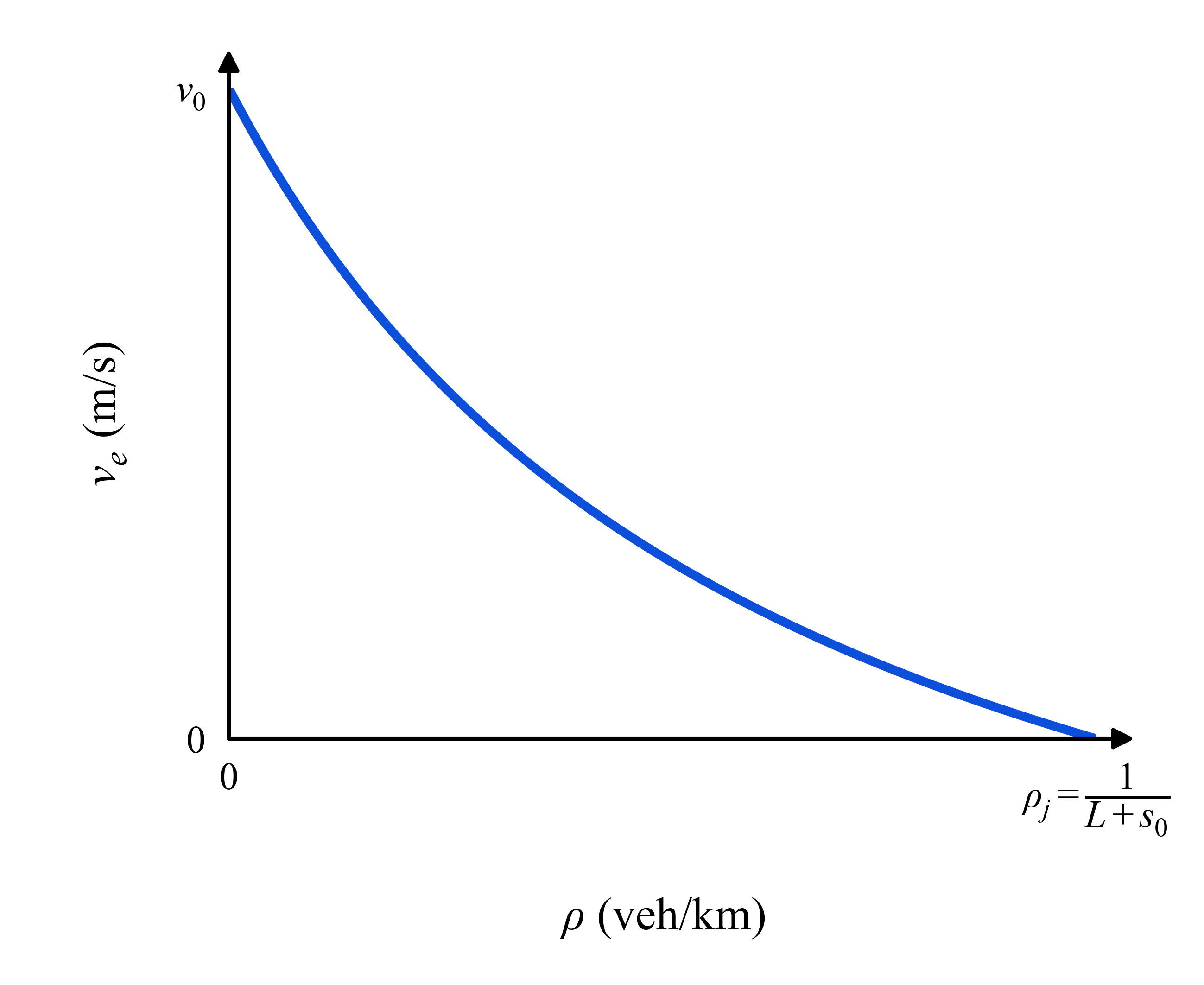}
    \hfill
    \includegraphics[width=0.48\textwidth]
    {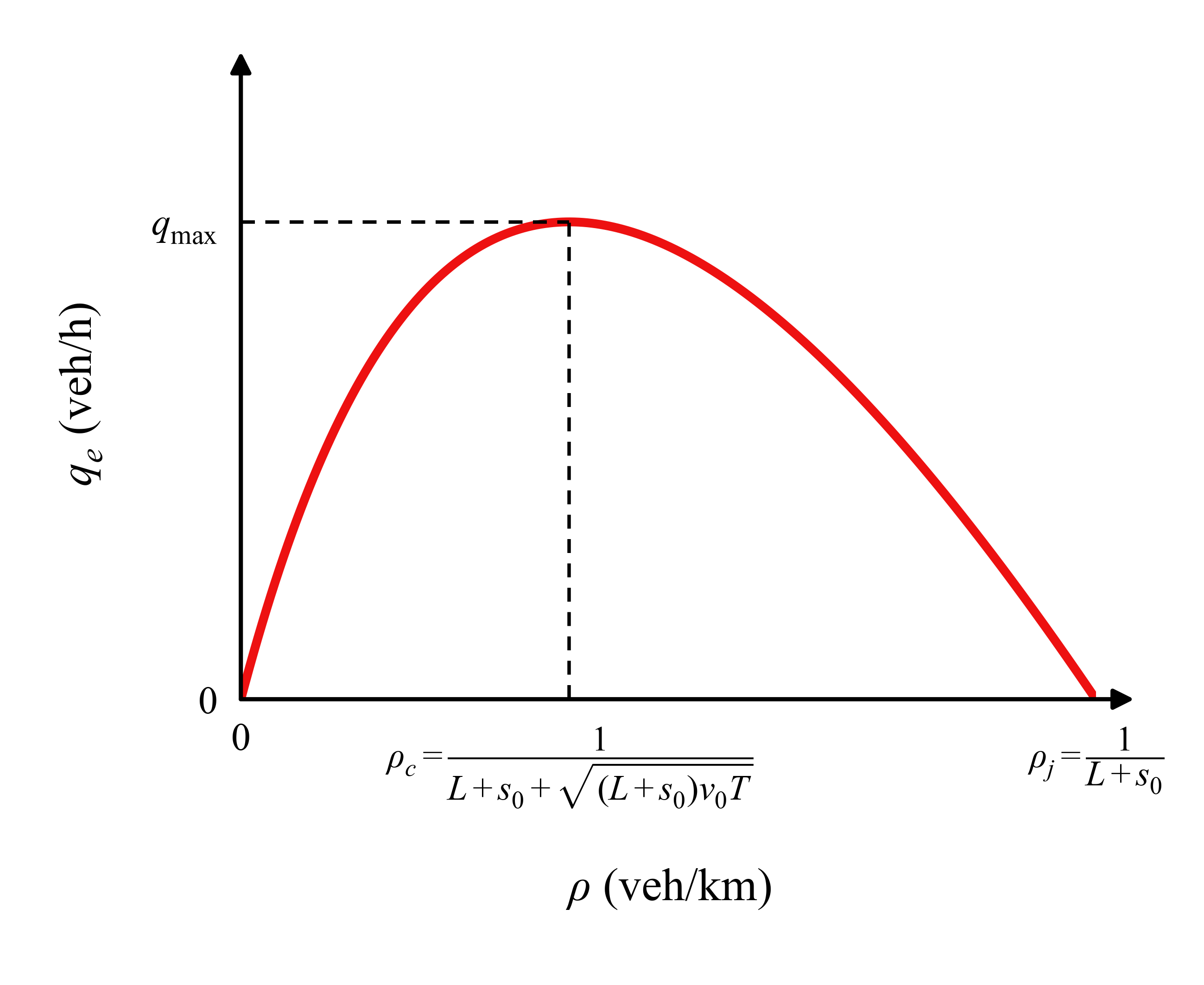}
    \caption{Model-implied fundamental diagrams of AERM:
    speed-density relationship (left) and
    flow-density relationship (right).}
    \label{fig:aerm_fd}
\end{figure*}

As shown in Fig.~\ref{fig:aerm_fd}, the equilibrium speed decreases monotonically with increasing density, approaching zero at the jam density. The corresponding flow-density curve exhibits a maximum at the critical density, separating the uncongested and congested equilibrium branches. These results show that AERM admits an explicit macroscopic fundamental diagram consistent with the expected equilibrium characteristics of traffic flow.

\subsubsection{Stability Analysis and Platoon Dynamics}
\label{sec:aerm_stability_platoon}

\paragraph{\textbf{Local and String Stability}}

Following the equilibrium analysis, the local and string stability of AERM are examined to assess its dynamic response to small perturbations and their propagation along vehicle platoons. For a homogeneous equilibrium state $(s^*,v^*)$ satisfying $\Delta v^*=0$ and $v^*=V_e(s^*)$, the unsaturated, zero-delay AERM is linearized following \citep{sun2018stability,sun2020relationship}. The equilibrium partial derivatives are:

\begin{equation}
    \begin{aligned}
        F_s &=
        \frac{\lambda v_0^2 T}
        {(s^*-s_0+v_0T)^2},\\
        F_v &= -\lambda,\\
        F_{\Delta v} &=
        k\frac{v^*}{s^*}
        \left(1+\frac{s_0}{s^*}\right),
    \end{aligned}
    \label{eq:aerm_partial_derivatives}
\end{equation}
where $F_s$, $F_v$, and $F_{\Delta v}$ denote the equilibrium derivatives with respect to spacing, follower speed, and relative speed, respectively. The linearized dynamics of a follower with a constant-speed leader are characterized by:

\begin{equation}
    \sigma^2
    +(F_{\Delta v}-F_v)\sigma
    +F_s=0.
    \label{eq:aerm_local_stability}
\end{equation}

For positive model parameters and physically consistent equilibria with $s^*>s_0$, AERM satisfies $F_s>0$ and
$F_{\Delta v}-F_v>0$. Both roots therefore have negative real parts, indicating local stability of the zero-delay
model. The equilibrium-restoration term provides the spacing-dependent response, while the relative-motion
term contributes additional damping to speed perturbations. The propagation of disturbances along a vehicle platoon
is further governed by the string stability condition:

\begin{equation}
    F_v^2
    -2F_vF_{\Delta v}
    -2F_s
    \geq 0.
    \label{eq:aerm_string_stability}
\end{equation}

Substituting the derivatives of AERM yields:

\begin{equation}
    \lambda
    +
    2k\frac{v^*}{s^*}
    \left(1+\frac{s_0}{s^*}\right)
    \geq
    \frac{2v_0^2T}
    {(s^*-s_0+v_0T)^2}.
    \label{eq:aerm_string_stability_condition}
\end{equation}

The local stability condition indicates that the positive spacing sensitivity and damping effect ensure stable equilibrium responses under physically consistent parameter settings. For string stability, the equilibrium restoration rate $\lambda$ and relative-speed response coefficient $k$ contribute to disturbance attenuation, whereas a steeper equilibrium speed-spacing relationship increases the tendency for disturbance amplification. In particular, the relative-speed sensitivity depends on the equilibrium speed and spacing, indicating that disturbance propagation varies with traffic conditions. Consequently, AERM can exhibit different string stability regimes depending on its calibrated parameters and equilibrium states. These characteristics are further examined through numerical platoon simulations.

\paragraph{\textbf{Platoon Disturbance Propagation}}

\begin{figure*}[!t]
    \centering
    \includegraphics[width=0.48\textwidth]
    {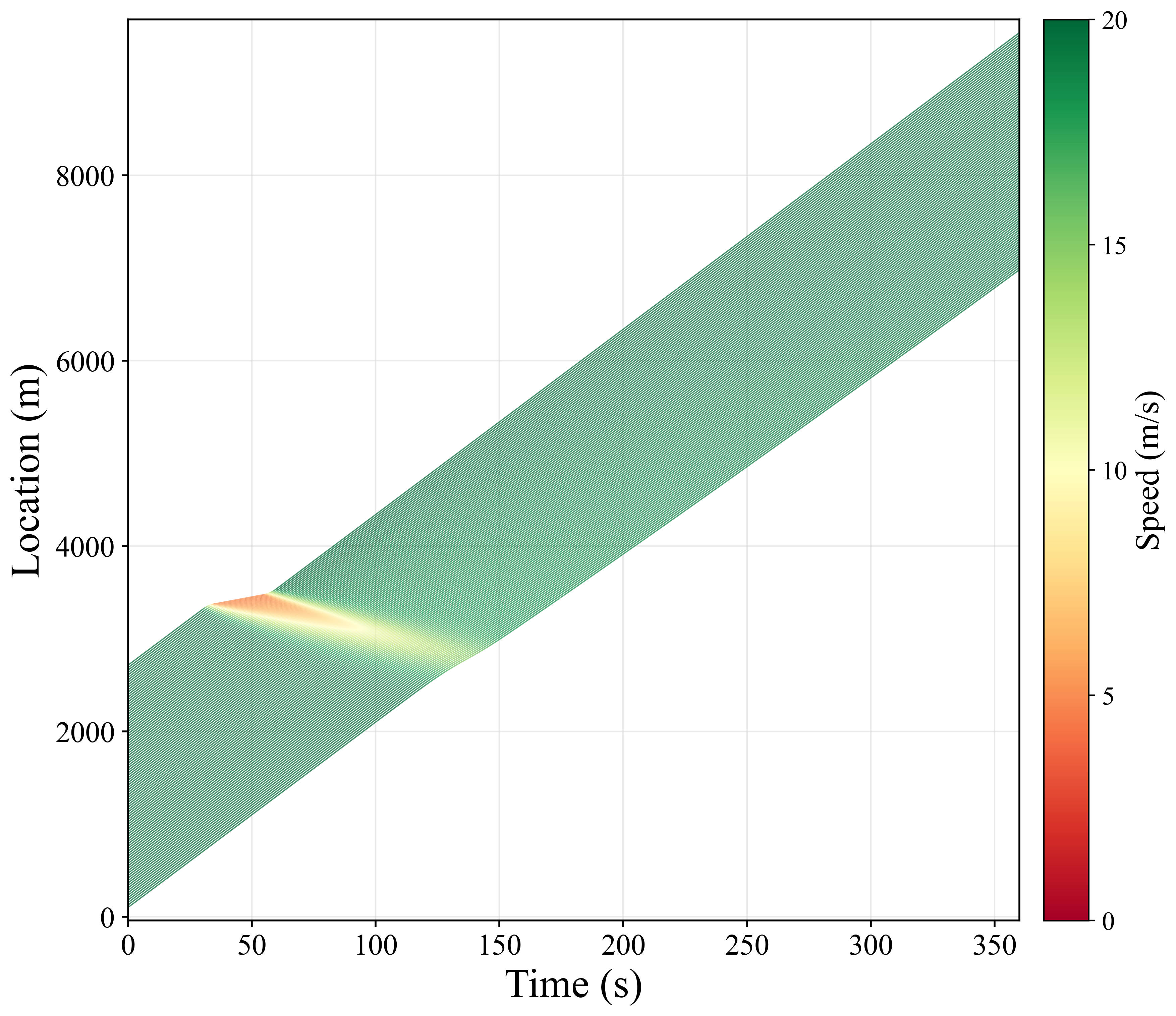}
    \hfill
    \includegraphics[width=0.48\textwidth]
    {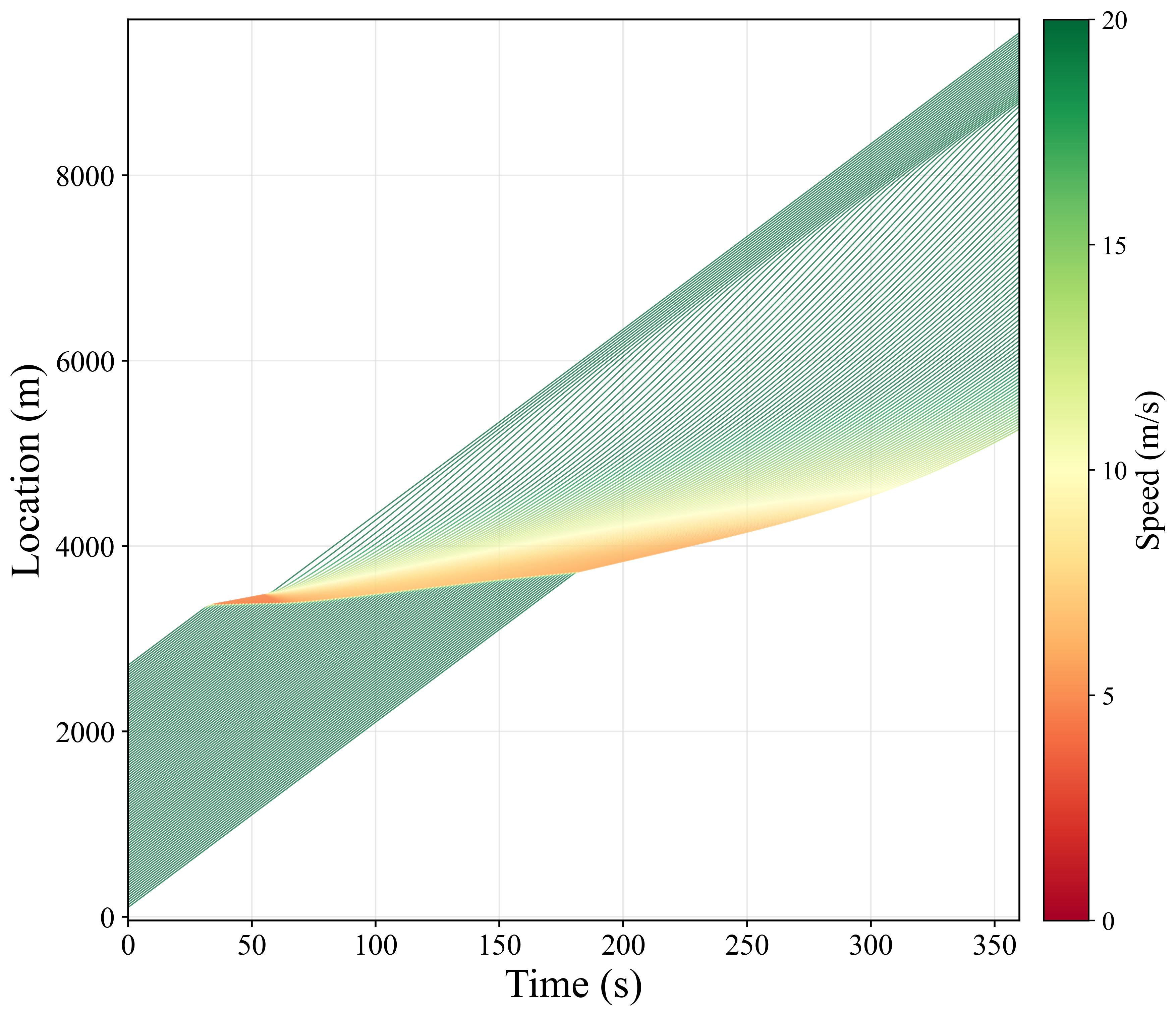}
    \caption{Spatiotemporal speed evolution under two
    AERM parameter configurations.
    (Left) Congestion dissipation: $k=1.05$,
    $\lambda=0.03$ s$^{-1}$.
    (Right) Congestion persistence: $k=0.25$,
    $\lambda=0.01$ s$^{-1}$.
    Shared parameters are $v_0=40$ m/s,
    $s_0=1.042$ m, and $T=0.509$ s.}
    \label{fig:aerm_platoon}
\end{figure*}

To examine the disturbance propagation characteristics of AERM, platoon simulations are conducted on a single-lane road with 100 vehicles, each 5 m long. All vehicles are
initially placed at a homogeneous equilibrium state with a speed of 20 m/s and a net spacing of 21.4 m. The simulation lasts 1200 s. A prescribed speed disturbance is applied to the leading vehicle: it maintains 20 m/s during the first 300 s, decelerates uniformly to 5 m/s
over the next 5 s, maintains this reduced speed for 20 s, and then accelerates uniformly back to 20 m/s over 5 s. The two simulation cases share identical initial conditions and leader disturbances. The resulting spatiotemporal speed distributions are presented in Fig.~\ref{fig:aerm_platoon}.

As shown in Fig.~\ref{fig:aerm_platoon}, the two parameter configurations produce distinct disturbance propagation patterns under identical initial conditions and leading-vehicle
perturbations. In the first configuration, the low-speed disturbance gradually attenuates following the leader's speed recovery, whereas in the second configuration, it persists and propagates upstream throughout the observation
period. These contrasting responses are consistent with the influence of relative-speed sensitivity and equilibrium-speed restoration on string stability. More importantly, they demonstrate that AERM can reproduce both disturbance dissipation and persistence through different parameter configurations while retaining the same analytical structure, highlighting its flexibility in representing different platoon-level traffic dynamics.

% ---------------------------------------------------------
\subsubsection{Microscopic Benchmarking}
\label{sec:aerm_microscopic_benchmarking}

Adopting the microscopic benchmarking framework proposed in \citep{NiWang2013Benchmarking}, we construct a controlled simulation scenario to compare the CF dynamics of the AERM and IDM. To guarantee identical test conditions, both following vehicles start from the same initial position and travel at a constant speed of $30$ m/s, while the original leader keeps stationary at $x_j=5000$ m. At $t=100$ s, a new leader moving at $24$ m/s cuts in, maintaining an identical net spacing of $40$ m for both followers. The leader subsequently executes a sequence of manoeuvres, where it decelerates at $3$ m/s$^2$ from $t=200$ s until full stop, accelerates at the same rate from $t=300$ s up to $36$ m/s, and finally decelerates again from $t=400$ s until full stop. The simulation terminates at $t=500$ s.

\begin{figure*}[!t]
    \centering
    \includegraphics[width=0.82\textwidth]
    {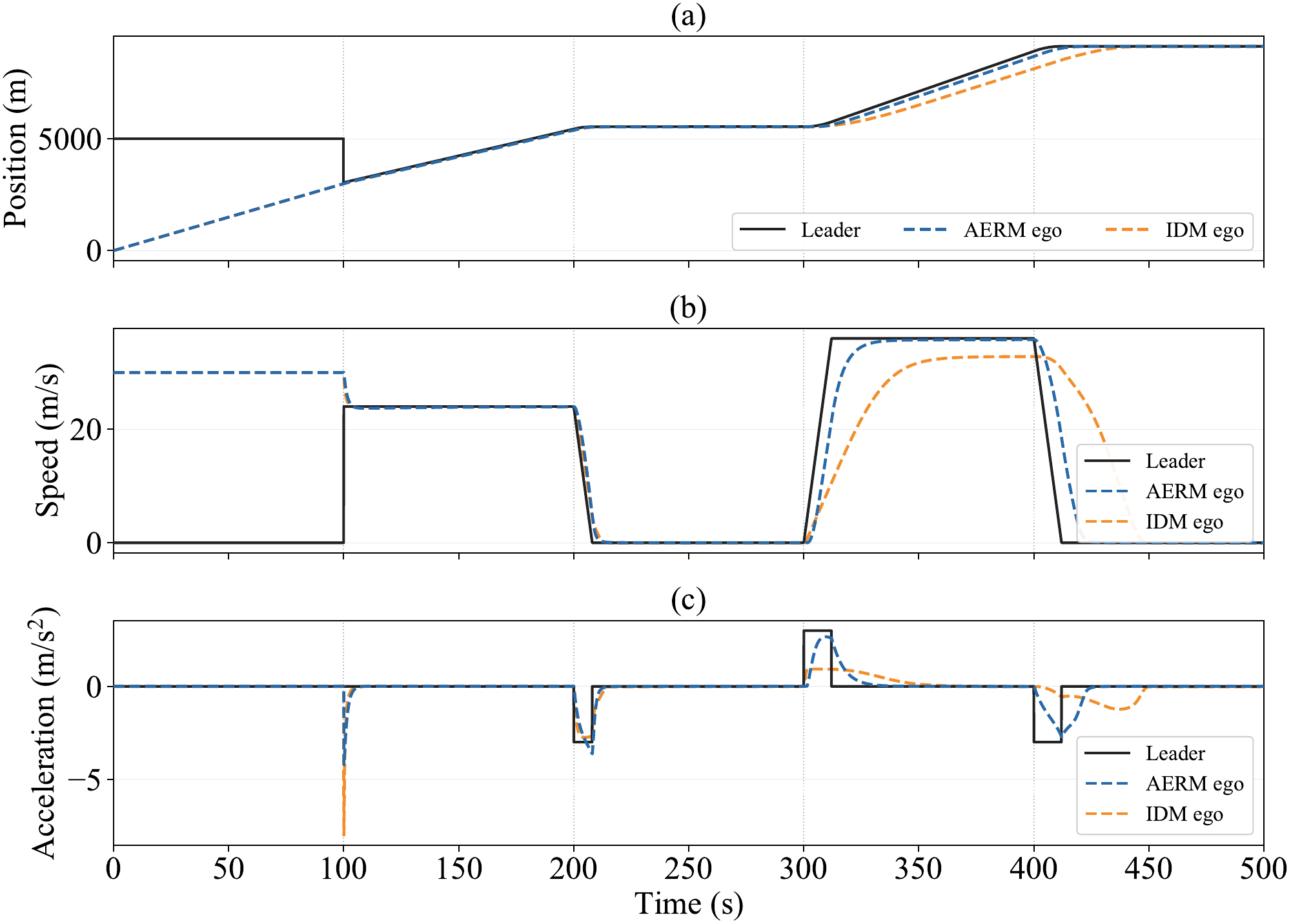}
    \caption{Microscopic benchmarking of AERM against IDM under a common prescribed leader trajectory: (a) vehicle trajectories, (b) longitudinal speed, and (c) longitudinal acceleration. Both AERM and IDM are calibrated using the Shanghai NDS dataset. The dataset-level calibrated parameters of AERM are reported in Table~\ref{tab:aerm_calibrated_parameters}, while the corresponding IDM parameters are $v_0=32.8118$, $T=1.3076$, $s_0=1.2977$, $a=0.9352$, and $b=0.8863$, with $\delta=4$.}
    \label{fig:aerm_idm_microscopic_benchmark}
\end{figure*}

Figure~\ref{fig:aerm_idm_microscopic_benchmark} demonstrates that both models generate reasonable car-following responses to the predefined leader manoeuvres, yet clear discrepancies arise in response magnitude and timing. AERM exhibits a faster acceleration during the recovery phase and converges toward the leader’s cruising speed of 36 m/s after restart, whereas IDM settles at a lower speed close to its desired speed $v_0=32.81$ m/s. Such distinctions are also visible in the acceleration time series and cumulative vehicle trajectories. Despite these divergent response characteristics, both models preserve positive net spacing throughout the entire simulation without collisions, verifying physically admissible car-following behavior under the tested leader manoeuvre sequence.

\begin{figure*}[!t]
    \centering
    \includegraphics[width=\textwidth]
    {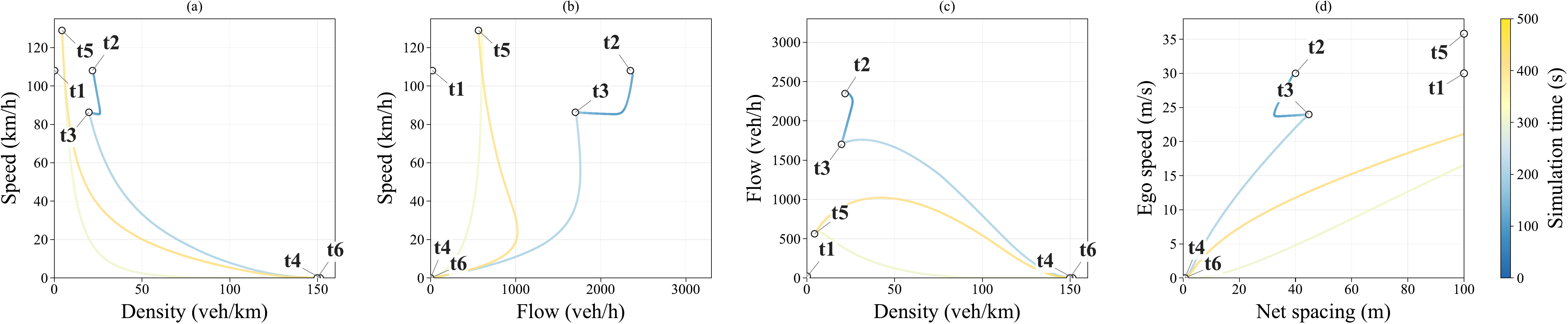}
    \caption{Time-ordered microscopic traffic-state
    trajectories of AERM: (a) speed-density, (b) speed-flow, (c) flow-density, and (d) speed-spacing. $t_1$--$t_6$ denote the initial cruising state, leader cut-in, first leader deceleration, leader acceleration, second leader deceleration, and final stationary state, respectively.} 
    \label{fig:aerm_microscopic_state_trajectories}
\end{figure*}

Figure~\ref{fig:aerm_microscopic_state_trajectories} further visualises the temporally ordered evolution of traffic states for AERM across four complementary state-space projections. The leader cut-in event at $t_2$ induces an abrupt discontinuity in the following spacing, followed by a gradual transition toward the leader’s steady cruising state. During the first deceleration phase of the leader $t_3$–$t_4$, the follower speed and flow drop toward zero, accompanied by an increase in traffic density. After the leader resumes motion at $t_4$, the follower accelerates progressively and adjusts its spacing, yielding a reverse transition across the state-space planes. The second deceleration command at $t_5$ then drives the follower toward the final stationary state at $t_6$. Notably, the resulting state trajectories display non-monotonic, loop-shaped patterns in multiple state-space representations, rather than collapsing onto a single equilibrium curve. This feature originates from the transient adaptation of follower speed to time-varying leader motion, and the instantaneous traffic state depends not merely on the current spacing, but also on historical motion and ongoing acceleration/deceleration processes. These observations confirm that AERM can capture time-dependent state transitions across distinct microscopic traffic regimes, beyond its static equilibrium relations.

% ---------------------------------------------------------
\section{Trajectory Calibration}
\label{sec:empirical_reconstruction}
To evaluate the applicability and trajectory calibration accuracy of the discovered AERM across various driving conditions, we compare its performance with seven established analytical CF models and four agent-based models. The evaluation uses three external datasets (NDS, NGSIM I-80, and Waymo), covering diverse driving environments and vehicle interactions. NGSIM I-80 provides a widely used benchmark for CF model evaluation, while NDS covers diverse drivers, road types, and driving conditions. Waymo further extends the evaluation to vehicle interactions recorded in autonomous driving environments. Two calibration protocols are employed to assess the reconstruction capability of AERM under dataset-level shared parameters and independently optimized trajectory-specific parameters.

\subsection{Dataset-Level Calibration}
\label{sec:aerm_common_parameter_results}

Under dataset-level calibration, each model uses a single parameter vector to reconstruct all CF trajectories within a dataset. The corresponding calibration results are reported in Table~\ref{tab:aerm_common_parameter_results}.

\afterpage{%
\begin{table*}[!ht]
\centering
\caption{Trajectory calibration performance with dataset-level parameters.}
\label{tab:aerm_common_parameter_results}
\footnotesize
\setlength{\tabcolsep}{5pt}
\renewcommand{\arraystretch}{1.15}

\begin{tabular*}{\textwidth}{
@{\extracolsep{\fill}}lrrrrrr@{}
}
\toprule
& \multicolumn{2}{c}{NDS}
& \multicolumn{2}{c}{NGSIM I-80}
& \multicolumn{2}{c}{Waymo}\\
\cmidrule(lr){2-3}
\cmidrule(lr){4-5}
\cmidrule(lr){6-7}
Model & Gap & Speed & Gap & Speed & Gap & Speed\\
\midrule
\multicolumn{7}{l}{\textit{Agent-based Models}} \\
SR-LLM-1
& 267.5789 & 13.2926
& 82.6832 & 5.1937
& 71.3132 & 7.8258 \\

SR-LLM-2
& 242.8610 & 11.8897
& 46.1716 & 3.6661
& 60.7075 & 6.7670 \\

SR-LLM-3
& 141.9852 & 5.7809
& 66.1628 & 3.8809
& 25.6086 & 3.4281\\

TR-Agent
& \textbf{8.8017} & 1.0145
& 5.1637 & 1.1944
& 5.8554 & 0.9539\\
\midrule
\multicolumn{7}{l}{\textit{Classical Models}} \\
GHR
& 10.0970 & \underline{0.9948}
& 5.6453 & 1.2050
& 5.5636 & \underline{0.8506}\\

Helly
& 9.3016 & 1.0749
& 5.3515 & 1.2242
& 5.9596 & 0.9303\\

Gipps
& 9.6389 & 1.0731
& 5.1731 & \underline{1.1270}
& \textbf{5.1695} & \textbf{0.8484}\\

Newell
& 11.0970 & 1.7086
& 6.1859 & 1.3207
& 7.7446 & 1.4549\\

OVM
& 12.7097 & 1.9593
& 6.3365 & 1.4418
& 22.1557 & 2.4185\\

FVDM
& 9.7358 & 1.0963
& 5.4598 & 1.2238
& 6.5767 & 1.0324\\

IDM
& 9.0403 & 0.9969
& \underline{5.1573} & 1.1347
& 5.5694 & 0.9021\\
\midrule
\textbf{AERM}
& \underline{8.9905} & \textbf{0.9911}
& \textbf{4.8045} & \textbf{1.1080}
& \underline{5.3708} & 0.9002\\

\bottomrule
\end{tabular*}

\par\smallskip
\begin{minipage}{0.98\textwidth}
\footnotesize

Gap and Speed denote RMSE in m and m/s, respectively.
Bold and underlined values indicate the lowest and
second-lowest errors in each column, respectively.
Collisions were observed for SR-LLM-2 on Shanghai NDS
(1.64\%) and Waymo (38.68\%), and for SR-LLM-1
(0.08\%), SR-LLM-2 (5.69\%), and SR-LLM-3 (0.04\%)
on NGSIM I-80. All other model--dataset combinations
had zero collisions.
\end{minipage}

\end{table*}
}

To further characterize the dataset-level calibration results, Table~\ref{tab:aerm_calibrated_parameters} reports the optimized parameter vectors of AERM for the three datasets. Although the analytical structure remains unchanged, the calibrated parameters vary across datasets, reflecting differences in driving characteristics and traffic conditions. In particular, the differences in $k$, $s_0$, and $\lambda$ indicate variations in the model's calibrated response characteristics across the three data sources.

\begin{table*}[!ht]
\centering
\caption{Dataset-level calibrated parameters of AERM.}
\label{tab:aerm_calibrated_parameters}
\footnotesize
\setlength{\tabcolsep}{5pt}
\renewcommand{\arraystretch}{1.15}

\begin{tabular*}{\textwidth}{
@{\extracolsep{\fill}}lccccc@{}
}
\toprule
Dataset & $k$ & $\lambda$ (s$^{-1}$) & $s_0$ (m)  
& $v_0$ (m/s) & $T$ (s) \\
\midrule
NDS
& 0.7656 & 0.0593 & 3.0973  & 40.0000 & 0.7099 \\

NGSIM I-80
& 1.2652 & 0.1538 & 3.7157  & 20.4885 & 1.1665 \\

Waymo
& 1.3937 & 0.0754 & 1.3763  & 40.0000 & 1.1676 \\
\bottomrule
\end{tabular*}
\end{table*}

\subsection{Trajectory-Specific Calibration}
\label{sec:aerm_trajectory_specific_results}

To further examine the fitting capability of AERM, each CF trajectory is independently calibrated while retaining the mathematical structure of the model. This protocol evaluates the reconstruction accuracy achievable by each analytical formulation when its parameters are individually optimized for the corresponding trajectory. The results are summarized in Table~\ref{tab:aerm_trajectory_specific_results}.

\begin{table*}[!ht]
\centering
\caption{Trajectory calibration performance with trajectory-specific parameters.}
\label{tab:aerm_trajectory_specific_results}
\footnotesize
\setlength{\tabcolsep}{5pt}
\renewcommand{\arraystretch}{1.15}

\begin{tabular*}{\textwidth}{
@{\extracolsep{\fill}}lrrrrrr@{}
}
\toprule
& \multicolumn{2}{c}{NDS}
& \multicolumn{2}{c}{NGSIM I-80}
& \multicolumn{2}{c}{Waymo}\\
\cmidrule(lr){2-3}
\cmidrule(lr){4-5}
\cmidrule(lr){6-7}
Model & Gap & Speed & Gap & Speed & Gap & Speed\\
\midrule
\multicolumn{7}{l}{\textit{Agent-based Models}} \\
SR-LLM-1
& 12.6180 & 1.3360
& 14.6429 & 1.6550
& 8.3390 & 2.0395 \\

SR-LLM-2
& 9.5288 & 1.1681
& 8.1314 & 1.5719
& 7.0855 & 1.7163 \\

SR-LLM-3
& 4.1166 & 0.8020
& 5.4065 & 1.0980
& 3.4018 & 0.9747 \\

TR-Agent
& 1.9191 & 0.5549
& 1.6390 & 0.9173
& 1.0972 & 0.4461\\
\midrule
\multicolumn{7}{l}{\textit{Classical Models}} \\
GHR
& 4.4980 & 0.7551
& 2.8763 & 1.0125
& 2.0739 & 0.5697\\

Helly
& 2.0142 & 0.5962
& 1.5750 & 0.8954
& 0.9933 & 0.4322\\

Gipps
& 1.7206 & 0.5289
& \underline{1.3725} & \underline{0.8658}
& \underline{0.6683} & 0.3551\\

Newell
& 2.0394 & 0.6002
& 1.7771 & 0.9102
& 1.0450 & 0.4665\\

OVM
& 2.1060 & 0.6162
& 1.8906 & 0.9504
& 1.3486 & 0.5488\\

FVDM
& 1.6517 & 0.5218
& 1.5542 & 0.8823
& 0.8341 & 0.3955\\

IDM
& \underline{1.5482} & \underline{0.4981}
& 1.4057 & 0.8678
& 0.6913 & \underline{0.3538}\\
\midrule
\textbf{AERM}
& \textbf{1.4071} & \textbf{0.4626}
& \textbf{1.2683} & \textbf{0.8454}
& \textbf{0.5231} & \textbf{0.3173}\\

\bottomrule
\end{tabular*}

\par\smallskip
\begin{minipage}{0.98\textwidth}
\footnotesize

Gap and Speed denote RMSE in m and m/s, respectively. Bold and underlined values indicate the lowest and second-lowest errors in each column, respectively. On Shanghai NDS and Waymo, collisions were observed only for SR-LLM-2, with rates of 8.76\% and 54.73\%, respectively. On NGSIM I-80, SR-LLM-1, SR-LLM-2, and SR-LLM-3 exhibited collision rates of 0.08\%, 27.28\%, and 0.04\%, respectively, while all other models recorded a collision rate of 0.0\%.
\end{minipage}

\end{table*} 

AERM achieves the lowest gap and speed calibration errors among all eleven models across the three external datasets, outperforming the strongest baselines in all six evaluation metrics. On NDS, AERM reduces gap and speed RMSEs by 9.11\% and 7.13\%, respectively, relative to IDM. On NGSIM I-80, the corresponding improvements reach 7.59\% and 2.36\% compared with Gipps. The advantage is particularly pronounced on Waymo, where 1,564 CF trajectories are evaluated. AERM achieves a spacing RMSE of 0.5231 m and a speed RMSE of 0.3173 m/s, representing reductions of 21.73\% relative to Gipps and 10.32\% relative to IDM, respectively. The consistent improvements demonstrate the strong trajectory-fitting capability of the discovered AERM structure. In particular, its substantial advantage on Waymo under trajectory-specific calibration contrasts with the common-parameter results, indicating that the analytical formulation can accommodate individual driving responses more accurately when its coefficients are independently optimized.

\begin{figure*}[!t]
    \centering
    \includegraphics[width=\textwidth, trim=0.8cm 0.8cm 0.5cm 0.3cm, clip]
    {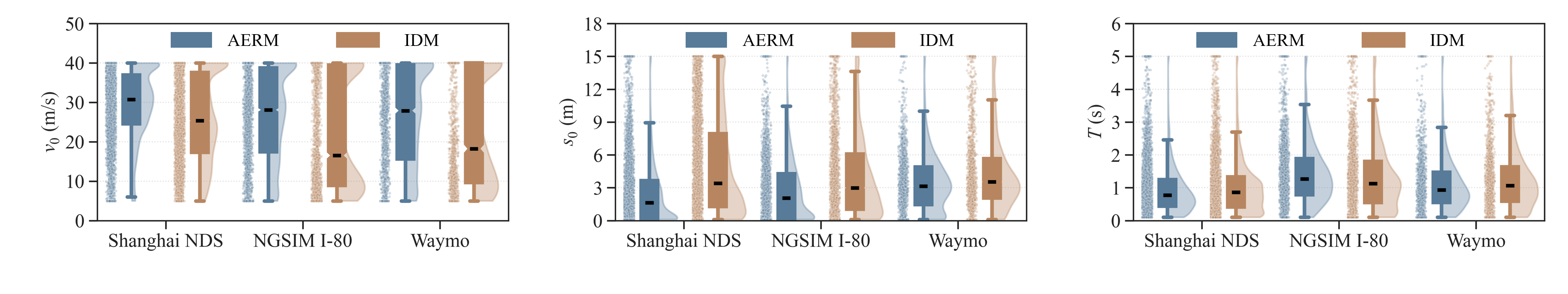}
    \caption{Distributions of trajectory-specific calibrated
    parameters $v_0$, $s_0$, and $T$ for AERM and IDM
    across the three external datasets.}
    \label{fig:parameter_distributions}
\end{figure*}

Figure~\ref{fig:parameter_distributions}
compares the trajectory-specific parameter distributions of AERM and IDM across the three external datasets. The calibrated $s_0$ values are predominantly on the order of a few meters, while $T$ is generally concentrated around 1 s, broadly consistent with typical parameter scales reported in empirical CF studies \citep{Treiber2000,Kesting2008}. Compared with IDM, AERM generally yields higher desired-speed estimates $v_0$ and smaller standstill spacing $s_0$, particularly on NDS and NGSIM I-80. The distributions of $T$ are broadly comparable between the two models, with some variations across datasets. Overall, both models produce physically plausible parameter estimates, while exhibiting distinct calibration patterns across heterogeneous driving conditions.

\begin{figure*}[!ht]
    \centering
    \includegraphics[
        width=\textwidth,
        trim=0cm 0cm 0cm 1.1cm,
        clip
    ]{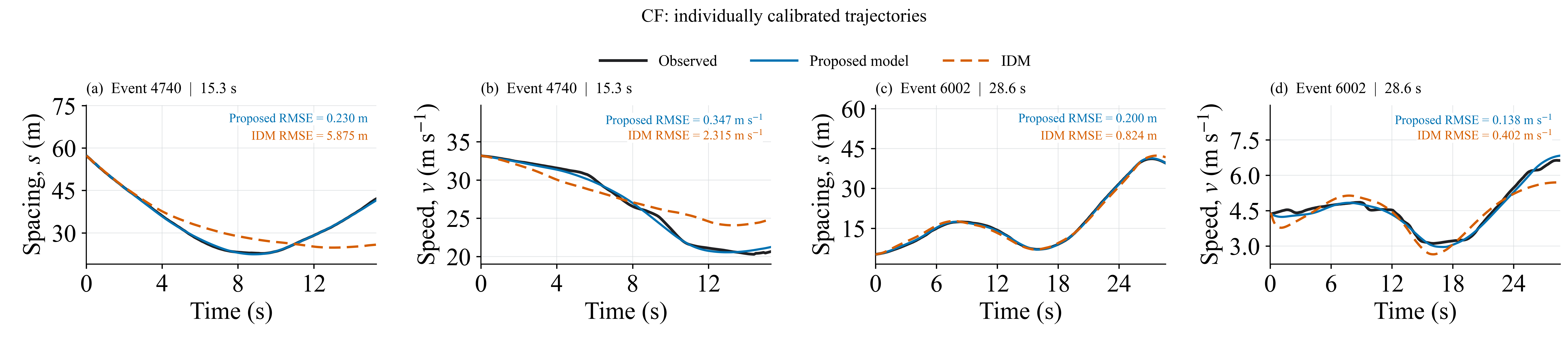}
    \caption{Comparison of observed and reconstructed CF trajectories using AERM and IDM on NDS.}
    \label{fig:aerm_shanghai_trajectory}
\end{figure*}

\begin{figure*}[!ht]
    \centering
    \includegraphics[
        width=\textwidth,
        trim=0cm 0cm 0cm 1.1cm,
        clip
    ]{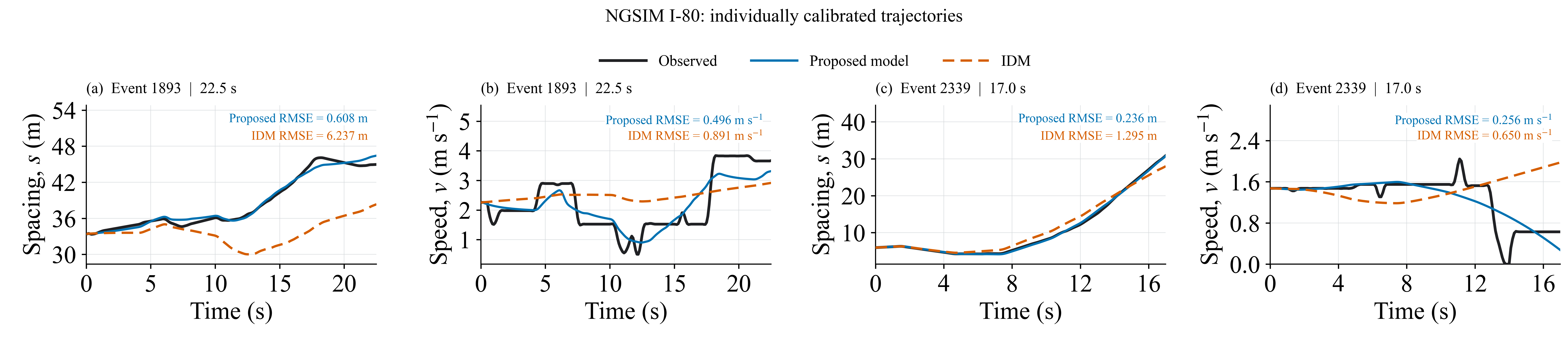}
    \caption{Comparison of observed and reconstructed CF trajectories using AERM and IDM on NGSIM I-80.}
    \label{fig:aerm_i80_trajectory}
\end{figure*}

\begin{figure*}[!ht]
    \centering
    \includegraphics[
        width=\textwidth,
        trim=0cm 0cm 0cm 1.1cm,
        clip
    ]{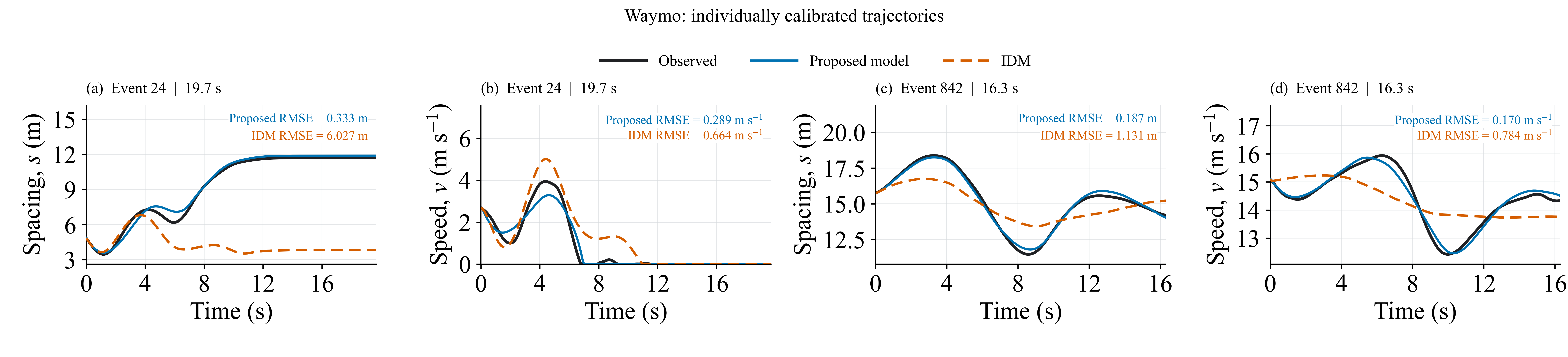}
    \caption{Comparison of observed and reconstructed CF trajectories using AERM and IDM on Waymo.}
    \label{fig:aerm_waymo_trajectory}
\end{figure*}

To illustrate the trajectory-level differences in calibration performance, representative CF trajectories from the three external datasets are compared between AERM and IDM. Both models are independently calibrated for each trajectory, and the reconstructed speed and spacing profiles are compared with the observations. The results are presented in Figs.~\ref{fig:aerm_shanghai_trajectory}--\ref{fig:aerm_waymo_trajectory}.

Figures~\ref{fig:aerm_shanghai_trajectory}--\ref{fig:aerm_waymo_trajectory} compare the observed and reconstructed follower-speed and spacing profiles for representative trajectories from the three external datasets. Across these examples, AERM reproduces the observed trajectories more closely than IDM, consistent with the lower trajectory-specific RMSEs reported in Table~\ref{tab:aerm_trajectory_specific_results}. These comparisons provide trajectory-level evidence of the fitting capability of the discovered AERM structure under independent parameter calibration.

\subsection{Ablation Study}
\label{sec:discovery_effectiveness}

An ablation study is conducted to examine the contributions of the knowledge-driven agentic search pathway, evolutionary search pathway, and physics-aware evaluation to empirical CF law discovery. Three configurations are considered: \emph{w/o Knowledge-driven agentic Pathway}, \emph{w/o Evolutionary Pathway}, and \emph{w/o Physical Constraints}. The first two remove the corresponding search pathways, while the third removes physical consistency constraints and the interpretability objective from candidate selection, retaining essential numerical safeguards.

All configurations employ NGSIM US-101, highD, and exiD for equation discovery. The resulting mathematical structures are subsequently evaluated on NGSIM I-80, which is excluded from the discovery process. Model coefficients are recalibrated under the same trajectory-specific GA protocol to compare the reconstruction capabilities of the discovered equations. The complete framework produces AERM, as presented in Eq.~\eqref{eq:aerm_model}. The mathematical structures obtained under the four configurations are summarized in Table~\ref{tab:ablation_equations}.

\begin{table*}[!ht]
\centering
\caption{Analytical CF equations discovered under different ablation configurations.}
\label{tab:ablation_equations}
\footnotesize
\setlength{\tabcolsep}{6pt}
\renewcommand{\arraystretch}{1.65}

\begin{tabularx}{\textwidth}{
@{}
>{\raggedright\arraybackslash}p{4.8cm}
>{\raggedright\arraybackslash}X
@{}
}
\toprule
\textbf{Configuration} &
\textbf{Discovered CF equation}\\
\midrule

\textbf{Full Framework (AERM)} &
$\displaystyle
\begin{aligned}
a={}&
\lambda\left[
\frac{v_0(s-s_0)}{s-s_0+v_0T}-v
\right]  
+k\frac{v}{s}
\left(1+\frac{s_0}{s}\right)\Delta v
\end{aligned}
$\\

%\midrule

w/o Knowledge-driven agentic Pathway &
$\displaystyle
a=
k_1\frac{\Delta v}{s}
+k_2s-\frac{k_3}{s}-k_4v+k_5
$\\

%\midrule

w/o Evolutionary Pathway &
$\displaystyle
\begin{aligned}
a={}&
\left(\frac{k_1}{s}-k_2v\right)\Delta v
+\lambda\left[
v_0\tanh\left(\beta(s-s_0)\right)-v
\right]
\end{aligned}
$\\

%\midrule

w/o Physical Constraints &
$\displaystyle
\begin{aligned}
a={}&
\lambda(v_0-v)
+\frac{\Delta v-\Delta v_0}
{s\left[
c_1|\Delta v|(c_2|v|)^m
\right]^n}
\end{aligned}
$\\

\bottomrule
\end{tabularx}

\par\smallskip
\begin{minipage}{0.98\textwidth}
\footnotesize
The equations are presented in parameterized form.
Coefficients are specific to each discovered structure.
\end{minipage}
\end{table*}

The structural differences reveal how each component affects the discovered CF law. Without the knowledge-driven agentic search pathway, the equation combines a relative-speed feedback term proportional to $\Delta v/s$ with separate spacing, inverse-spacing, speed-damping, and constant terms. Without evolutionary search, the equation retains a two-component structure comprising state-dependent relative-speed feedback and a $\tanh$-based equilibrium-speed restoration term. Its overall structure resembles AERM, although the functional forms differ. Without physical constraints, the discovered equation contains a more complex nonlinear response involving nested powers and absolute-value terms. Its raw expression is singular at $\Delta v=0$, requiring numerical safeguards during evaluation. In comparison, AERM combines state-dependent relative-speed response with an explicit rational equilibrium-speed relationship, providing a more structured representation of longitudinal vehicle interactions. The calibration performance of the four configurations is summarized in Table~\ref{tab:ablation_results}.

\begin{table*}[!ht]
\centering
\caption{Ablation results of the proposed CF law discovery framework on NGSIM I-80.}
\label{tab:ablation_results}
\footnotesize
\setlength{\tabcolsep}{6pt}
\renewcommand{\arraystretch}{1.15}

\begin{tabular*}{\textwidth}{
@{\extracolsep{\fill}}lcc@{}
}
\toprule
Configuration &
spacing RMSE (m) $\downarrow$ &
Speed RMSE (m/s) $\downarrow$ \\
\midrule

\textbf{Full Framework}
& \textbf{1.2683}
& \textbf{0.8454} \\

w/o Knowledge-driven agentic Pathway
& 4.4877
& 1.0800 \\

w/o Evolutionary Pathway
& 4.5314
& 1.0812 \\

w/o Physical Constraints
& 8.7521
& 1.4986 \\

\bottomrule
\end{tabular*}
\end{table*}

As shown in Table~\ref{tab:ablation_results}, the complete framework achieves the lowest gap and speed RMSEs among all configurations. \textbf{Compared with the variants without the knowledge-driven agentic and evolutionary search pathways}, AERM reduces spacing RMSE by 71.74\% and 72.01\%, respectively, with corresponding speed RMSE reductions of 21.72\% and 21.81\%. These results suggest that the complementary search pathways contribute to the discovery of more accurate analytical relationships. Removing physical constraints causes the largest deterioration in calibration performance. Spacing RMSE increases from 1.2683 m to 8.7521 m, while speed RMSE increases from 0.8454 m/s to 1.4986 m/s. The resulting formula also exhibits greater structural complexity and less straightforward physical interpretation. 

Overall, the ablation results demonstrate the superior calibration performance of the complete framework in this comparison. The differences among the discovered equations further indicate that complementary search and physics-aware evaluation influence both the empirical accuracy and mathematical structure of the resulting CF laws.

\section{Discussion}
This paper presents an agentic discovery workflow and the resulting AERM. The discussion interprets model behavior, compares it against existing families of CF formulations, reflects on evaluation boundaries, explains the role of each component in the discovery pipeline, and highlights remaining limitations for future research.

The interpretive value of AERM originates from two structural features uncovered through agent‑guided symbolic search. First, the equilibrium‑restoration term delivers an explicit closed‑form mapping from net spacing to equilibrium speed. Second, the relative‑speed feedback adopts state‑dependent sensitivity instead of constant gain. Combined, these two components support full analytical derivation of equilibrium relationships, fundamental‑diagram expressions, as well as local and string stability criteria. Platoon‑level simulation further demonstrates that a single fixed functional form can generate qualitatively distinct traffic phenomena, as with different parameter sets, AERM reproduces both disturbance dissipation and persistent backward‑propagating congestion, without modifying model structure. This shows the model can represent heterogeneous platoon dynamics purely via parameter heterogeneity.

Viewed in the context of classical CF theory, AERM brings together complementary behavioral principles that have been represented in different model formulations. Its equilibrium-speed restoration reflects the optimal-velocity principle embodied in OVM, while its relative-speed response is related to the stimulus--response and velocity-difference mechanisms represented by GHR and FVDM. The spacing-dependent equilibrium relationship also shares physical characteristics with Newell-type formulations, while the parameters $s_0$, $T$, and $v_0$ retain familiar interpretations in desired-spacing models such as IDM. These established formulations provided behavioral references during the agentic discovery process, in which their underlying principles were explored alongside empirical reconstruction diagnostics. The resulting AERM does not reproduce any single classical formulation, but expresses these complementary behavioral principles through a rational equilibrium-speed function and a state-dependent relative-speed gain. 

Structurally, AERM sits within the well‑established optimal‑velocity model family but resolves several practical drawbacks of OVM and FVDM. Unlike OVM, microscopic dynamics and macroscopic fundamental diagram are internally consistent without additional post‑hoc adjustment. Different from FVDM, relative‑speed response strength varies with driving state, matching the expectation that human or automated‑vehicle anticipatory responses strengthen under short following gaps. Nevertheless, AERM also introduces new constraints, as the prescribed rational form of the equilibrium-speed function may limit its flexibility in reproducing diverse empirical fundamental-diagram shapes.

When interpreting calibration results across external datasets, it is critical to distinguish between trajectory‑specific recalibration performance and cross‑dataset zero‑shot parameter transfer. The substantial RMSE reduction observed under trajectory‑specific calibration demonstrates the expressive power of AERM’s functional form. Given per‑episode parameter degrees‑of‑freedom, AERM is capable of fitting diverse longitudinal response patterns from both human‑driven and automated‑vehicle trajectories. However, such fitting performance should not be conflated with out‑of‑domain predictive generalization. In all external evaluations in this work, every competing baseline including AERM is calibrated using data from each target dataset. Parameters obtained from discovery datasets (NGSIM US‑101, highD, exiD) are not directly reused for prediction on external datasets. The reported accuracy gains mainly reflect the capacity of the mathematical structure to reproduce observed car‑following trajectories, rather than validating the transferability of calibrated parameter sets across traffic environments.

Ablation studies further illustrate the necessity of multiple components within our knowledge-driven agentic framework. Removing either the knowledge-driven agentic search pathway or the evolutionary symbolic search pathway substantially degrades reconstruction accuracy, confirming that the two search branches play complementary roles in exploring model structures. Disabling physical‑consistency and interpretability objectives produces highly complex mathematical expressions with poor empirical performance and numerical pathologies. This result warns that optimizing purely for trajectory‑fitting error is insufficient for discovering practically useful traffic‑flow laws, and physical plausibility constraints are indispensable even for data‑driven scientific discovery pipelines. %It should also be noted that our reported AERM originates from one complete execution of the closed‑loop discovery workflow. Symbolic search carries inherent stochasticity. While ablation shows that removing the diagnosis‑guided branch prevents recovery of the AERM‑like structure, repeated multi‑run experiments are required in future to statistically quantify how much the diagnosis‑agent module raises the probability of discovering physically meaningful closed‑form car‑following equations.

Several additional limitations remain to be addressed in follow‑up research. First, the current study focuses exclusively on longitudinal CF behavior, while lane‑changing and multi‑lane interaction are not considered. Second, we mostly focus on freeway trajectory datasets in this study, as generalization toward urban‑road car‑following scenarios requires further empirical examination.

\section{Conclusions}
\label{sec:conclusions}
In this study, we developed a knowledge-driven agentic framework for discovering interpretable CF laws from empirical trajectory data. The framework equips the reasoning agent with classical CF formulations, their behavioral interpretations, and fundamental traffic flow principles as domain knowledge priors. The agent combines this knowledge with empirical reconstruction diagnostics to formulate and refine structural hypotheses. Complementary evolutionary search explores alternative mathematical expressions, while physical consistency constraints and multi-objective evaluation guide the selection of interpretable and empirically supported CF laws.

The framework discovered the Adaptive Equilibrium Response Model (AERM), which combines spacing-dependent equilibrium-speed restoration with state-dependent relative-speed response, following the optimal-velocity principle. Its analytical structure admits explicit equilibrium relationships and fundamental diagrams, together with local and string stability conditions. Platoon simulations further show that different parameter configurations can reproduce congestion dissipation and persistence within the same mathematical structure. Under dataset-level calibration, AERM achieves the lowest speed RMSE on NDS and the lowest spacing and speed RMSEs on NGSIM I-80, while TR-Agent performs best in spacing reconstruction on NDS and Gipps achieves the best performance on Waymo. Under trajectory-specific calibration, AERM outperforms all eleven reference models across the three datasets, reducing spacing RMSE by 9.11\%, 7.59\%, and 21.73\%, respectively, relative to the strongest baselines. 
Overall, this study demonstrates the potential of a knowledge-driven agentic framework to discover analytical CF laws that combine empirical calibration performance with interpretable traffic-flow properties. %The findings provide a basis for further investigation of transferable and physically meaningful CF models across heterogeneous driving environments.

\section*{Replication and Data Sharing}

The source code for this work will be made publicly available upon acceptance. Data-sharing arrangements will be provided upon reasonable request, subject to institutional and privacy restrictions.

\section*{Declaration of Competing Interest}

The authors declare that they have no known competing financial interests or personal relationships that could have influenced the work reported in this paper.

\section*{Acknowledgement}

The authors would like to extend sincere thanks to Prof. Daiheng Ni at the University of Massachusetts Amherst for many insightful comments and suggestions. This research is partially supported by the National Natural Science Foundation of China (Nos. 52125208, 52402374), the Fundamental and Interdisciplinary Disciplines Breakthrough Plan of the Ministry of Education of China (No. JYB2025XDXM123), and the Fundamental Research Funds for the Central Universities.

\section*{Appendix}

The governing equations of the reference CF models are summarized in Table~\ref{tab:baseline_formulations}, with their parameter definitions provided in Table~\ref{tab:baseline_parameters}. Here, $v$ and $v_{\mathrm{L}}$ denote the follower and leader speeds, respectively, $s$ is the net spacing, and $\Delta v=v_{\mathrm{L}}-v$. Gipps is expressed in terms of its equivalent acceleration, while Newell adopts an acceleration-based relaxation formulation.

\begingroup
\footnotesize
\setlength{\tabcolsep}{12pt}
\renewcommand{\arraystretch}{2.8}

\begin{longtable}{@{}p{1.9cm}
    >{\raggedright\arraybackslash}
    p{\dimexpr\linewidth-1.9cm-2\tabcolsep\relax}@{}}

    \caption{Governing formulations of the reference
    CF models.}
    \label{tab:baseline_formulations}\\
    
    \toprule
    \textbf{Model} &
    \textbf{Reference governing formulation}\\
    \midrule
    \endfirsthead
    
    \toprule
    \textbf{Model} &
    \textbf{Reference governing formulation}\\
    \midrule
    \endhead
    
    \bottomrule
    \endfoot
        SR-LLM-1 &
        $\displaystyle
        a=\alpha\left[
        \left(c-\frac{v}{v_0}\right)^2-1
        \right]
        $
        \\[3pt]
        
        SR-LLM-2 &
        $\displaystyle
        a=b+
        \frac{b}{
        2c-\dfrac{v\Delta v}
        {2\sqrt{\alpha b}\,s}+1
        }
        -\alpha\sqrt{\frac{v}{v_0}}
        $
        \\[3pt]
        
        SR-LLM-3 &
        $\displaystyle
        a=\alpha\left[
        c\left(c-\frac{v}{v_0}\right)^2
        -\frac{v}{v_0}-1
        \right]
        +\frac{v\Delta v}{s}
        $
        \\[3pt]

        TR-Agent &
        $\displaystyle
        \begin{aligned}
        a&=(1-\mathrm{Sig})a_{\mathrm{free}}
        +\mathrm{Sig}\,a_{\mathrm{inter}},
        \quad
        \mathrm{Sig}=\frac{1}{1+\exp(\beta\Delta v)},\\[4pt]
        a_{\mathrm{free}}
        &=a_{\max}\left[1-\tanh\left(
        \beta\left(\frac{v}{v_0}-1\right)\right)\right],
        \quad
        a_{\mathrm{inter}}
        =-a_{\max}[1-\tanh(\beta Z)],\\[4pt]
        Z&=\frac{s-s^*(v,\Delta v)}
        {\max\{s^*(v,\Delta v),1\}},
        \quad
        s^*(v,\Delta v)
        =s_0+\max\left\{0,\,
        vT-\frac{v\Delta v}{2\sqrt{a_{\max}b}}
        \right\}.
        \end{aligned}
        $
        \\[3pt]

        GHR &
        $\displaystyle
        a
        =
        c
        \frac{v^{m}}
        {s^{l}}
        \Delta v
        $
        \\[3pt]

        Helly &
        $\displaystyle
        a
        =
        \alpha \Delta v
        +
        \gamma
        \left(
        s-s_0-Tv
        \right)
        $
        \\[3pt]

        Gipps &
        $\displaystyle
        \begin{aligned}
        a_{\mathrm G}
        &=\frac{\min\{v_{\mathrm{free}},v_{\mathrm{safe}}\}-v}
        {T_{\mathrm G}},
        \quad
        v_{\mathrm{free}}
        =v+2.5a_0T_{\mathrm G}
        \left(1-\frac{v}{v_0}\right)
        \sqrt{0.025+\frac{v}{v_0}},\\[4pt]
        v_{\mathrm{safe}}
        &=-b_fT_{\mathrm G}
        +\sqrt{
        b_f^2T_{\mathrm G}^2+
        b_f\left(
        2s-vT_{\mathrm G}
        +\frac{v_{\mathrm L}^2}{b_{\mathrm L}}
        \right)
        }.
        \end{aligned}
        $
        \\[3pt]

        Newell &
        $\displaystyle
        a_{\mathrm N}
        =k_{\mathrm N}\left[V_{\mathrm N}(s)-v\right],
        \quad
        V_{\mathrm N}(s)
        =\min\left\{
        v_0,\frac{s-s_0}{T_{\mathrm N}}
        \right\}
        $
        \\[3pt]
        
        OVM &
        $\displaystyle
        a_n
        =
        \alpha
        \left[
        V(s_n)-v_n
        \right],\quad
        V(s)
        =
        \frac{v_{\max}}{2}
        \left[
        \tanh
        \left(
        \frac{s-s_c}{\Delta s}
        \right)
        +
        \tanh
        \left(
        \frac{s_c}{\Delta s}
        \right)
        \right].
        $
        \\[3pt]

        FVDM &
        $\displaystyle
        a_n
        =
        \kappa
        \left[
        V(s_n)-v_n
        \right]
        +
        \lambda\Delta v_n,\quad
        V(s)
        =
        V_1
        +
        V_2
        \tanh
        \left[
        C_1(s-l_c)-C_2
        \right].
        $
        \\[3pt]

        IDM &
        $\displaystyle
        a
        =
        a_0
        \left[
        1
        -
        \left(
        \frac{v}{v_0}
        \right)^4
        -
        \left(
        \frac{s^*}{s}
        \right)^2
        \right],\quad
        s^*
        =
        s_0
        +
        \max\left\{0,\,
        vT
        -
        \frac{
        v\Delta v
        }{
        2\sqrt{a_0b}
        }
        \right\}.
        $
        \\
    
\end{longtable}
\endgroup

\begingroup
\footnotesize
\setlength{\tabcolsep}{6pt}
\renewcommand{\arraystretch}{1.18}
\setlength{\LTleft}{0pt}
\setlength{\LTright}{0pt}

\begin{longtable}{@{}
    p{2.0cm}
    p{2.0cm}
    >{\raggedright\arraybackslash}
    p{\dimexpr\textwidth-4.0cm-4\tabcolsep\relax}
    @{}}

\caption{Parameter definitions of the reference CF models.}
\label{tab:baseline_parameters}\\

\toprule
\textbf{Model} &
\textbf{Parameter} &
\textbf{Physical interpretation}\\
\midrule
\endfirsthead

\toprule
\textbf{Model} &
\textbf{Parameter} &
\textbf{Physical interpretation}\\
\midrule
\endhead

\bottomrule
\endfoot
SR-LLM-1
& $\alpha$ & Acceleration scale\\
& $c$ & Dimensionless nonlinear response coefficient\\
& $v_0$ & Desired speed\\
\addlinespace[4pt]

SR-LLM-2
& $\alpha$ & Acceleration-scale coefficient\\
& $b$ & Acceleration-scale coefficient\\
& $c$ & Dimensionless response coefficient\\
& $v_0$ & Desired speed\\
\addlinespace[4pt]

SR-LLM-3
& $\alpha$ & Acceleration scale\\
& $c$ & Dimensionless nonlinear response coefficient\\
& $v_0$ & Desired speed\\
\addlinespace[4pt]

TR-Agent
& $a_{\max}$ & Acceleration scale\\
& $b$ & Comfortable deceleration\\
& $v_0$ & Desired free-flow speed\\
& $s_0$ & Standstill spacing\\
& $T$ & Desired time gap\\
& $\beta$ & Nonlinear response sensitivity \\

GHR
& $c$ & Response sensitivity coefficient\\
& $m$ & Speed exponent\\
& $l$ & Spacing exponent\\
\addlinespace[4pt]

Helly
& $\alpha$ & Relative-speed response coefficient\\
& $\gamma$ & Spacing-error response coefficient\\
& $s_0$ & Standstill spacing\\
& $T$ & Desired time gap\\
\addlinespace[4pt]

Gipps
& $a_0$ & Acceleration parameter\\
& $v_0$ & Desired speed\\
& $T_{\mathrm G}$ & Update interval\\
& $b_f$ & Follower braking deceleration magnitude\\
& $b_{\mathrm L}$ & Leader braking deceleration magnitude\\
\addlinespace[4pt]

Newell
& $k_{\mathrm N}$ & Relaxation rate\\
& $v_0$ & Free-flow speed\\
& $s_0$ & Standstill spacing\\
& $T_{\mathrm N}$ & Time-gap parameter\\
\addlinespace[4pt]

OVM
& $\alpha$ & Relaxation rate\\
& $v_{\max}$ & Speed scale\\
& $s_c$ & Characteristic spacing\\
& $\Delta s$ & Spacing transition scale\\
\addlinespace[4pt]

FVDM
& $\kappa$ & Optimal-velocity relaxation rate\\
& $\lambda$ & Relative-speed response coefficient\\
& $V_1$ & Optimal-velocity offset\\
& $V_2$ & Optimal-velocity amplitude\\
& $C_1$ & Spacing sensitivity coefficient\\
& $C_2$ & Optimal-velocity shape coefficient\\
& $l_c$ & Characteristic spacing offset\\
\addlinespace[4pt]

IDM
& $a_0$ & Maximum acceleration\\
& $b$ & Comfortable deceleration\\
& $v_0$ & Desired free-flow speed\\
& $s_0$ & Standstill spacing\\
& $T$ & Desired time headway\\
\addlinespace[4pt]

\end{longtable}
\endgroup

\clearpage
\bibliographystyle{elsarticle-harv-no-links}
\bibliography{references}

\end{document}